\pdfoutput=1
\documentclass[preprint]{elsarticle}
\usepackage[a4paper, total={6.1in,9.4in}]{geometry}
\usepackage{lineno}
\usepackage[utf8]{inputenc} % allow utf-8 input
\usepackage[T1]{fontenc}    % use 8-bit T1 fonts
\usepackage{color}
\usepackage{hyperref}       % hyperlinks
\usepackage{url}            % simple URL typesetting
\usepackage{makecell}
\usepackage{amsmath,amsfonts,amsthm,bm} %math
\usepackage{booktabs}       % professional-quality tables
\usepackage{amsfonts}       % blackboard math symbols
\usepackage{nicefrac}       % compact symbols for 1/2, etc.
\usepackage{lipsum}
\usepackage{ragged2e}
\usepackage{setspace}
\usepackage{array}
\usepackage{lscape}
\usepackage{multirow}
\usepackage{microtype}
\usepackage{natbib}
\usepackage[font=small,labelfont=bf]{caption}
\usepackage{wrapfig}
\newcommand{\cmark}{\ding{51}}%
\newcommand{\xmark}{\ding{55}}%
\newcommand{\han}[1]{{\color{black}{#1}}}
\journal{Neural Network}
\begin{document}
\begin{frontmatter}
\title{Deformation-invariant Neural Network and Its Applications in Distorted Image Restoration and Analysis}
\author{Han Zhang\fnref{fn1,fn2}}
\ead{hzhang863-c@my.cityu.edu.hk}
\author{Qiguang Chen\fnref{fn3}}
\ead{qgchen@math.cuhk.edu.hk}
\author{Lok Ming Lui\fnref{fn3}\corref{cor1}}
\ead{lmlui@math.cuhk.edu.hk}
\cortext[cor1]{Corresponding author}
\fntext[fn1]{City University of Hong Kong, Kowloon, Hong Kong, China}
\fntext[fn2]{Hong Kong Center for Cerebro-Cardiovascular Health Engineering, Sha Tin, Hong Kong, China}
\fntext[fn3]{Chinese University of Hong Kong, Sha Tin, Hong Kong, China}

\begin{abstract}
Images degraded by geometric distortions pose a significant challenge to imaging and computer vision tasks such as object recognition. Deep learning-based imaging models usually fail to give accurate performance for geometrically distorted images. In this paper, we propose the deformation-invariant neural network (DINN), a framework to address the problem of imaging tasks for geometrically distorted images. The DINN outputs consistent latent features for images that are geometrically distorted but represent the same underlying object or scene. The idea of DINN is to incorporate a simple component, called the quasiconformal transformer network (QCTN), into other existing deep networks for imaging tasks. The QCTN is a deep neural network that outputs a quasiconformal map, which can be used to transform a geometrically distorted image into an improved version that is closer to the distribution of natural or good images. It first outputs a Beltrami coefficient, which measures the quasiconformality of the output deformation map. By controlling the Beltrami coefficient, the local geometric distortion under the quasiconformal mapping can be controlled. The QCTN is lightweight and simple, which can be readily integrated into other existing deep neural networks to enhance their performance. Leveraging our framework, we have developed an image classification network that achieves accurate classification of distorted images. Our proposed framework has been applied to restore geometrically distorted images by atmospheric turbulence and water turbulence. DINN outperforms existing GAN-based restoration methods under these scenarios, demonstrating the effectiveness of the proposed framework. Additionally, we apply our proposed framework to the 1-1 verification of human face images under atmospheric turbulence and achieve satisfactory performance, further demonstrating the efficacy of our approach.
\end{abstract}
\end{frontmatter}
% \linenumbers
\section{Introduction}

Deep learning methods have made significant strides in the field of imaging and computer vision, allowing us to achieve remarkable results in tasks like image restoration, object recognition, and classification. However, when it comes to degraded images, deep learning methods can face significant challenges. One such category of degraded images is those that are corrupted by geometric distortion, such as atmospheric turbulence or water turbulence. The use of deep learning methods may fail to produce accurate results for such images. For example, in the facial recognition task for images obtained by long-range cameras, the facial structure in the images is often geometrically distorted due to atmospheric turbulence, causing classical classification networks to provide incorrect results~\cite{lau2020atfacegan}. One intuitive approach for solving this problem is to add distorted images to the downstream classification networks for fine-tuning. However, this approach can be expensive due to the typically large size of the downstream network. Additionally, the introduction of extra variance in the data distribution caused by the distorted images may potentially degrade the performance of the tuned neural network. This challenge motivates the development of a framework that can effectively deal with geometrically distorted images, enabling deep learning methods to achieve accurate and reliable results even in challenging conditions.

There are two possible approaches to address this problem. One approach is to integrate a physical model that describes the geometric distortion. However, finding an appropriate physical model to describe the different types of geometric deformations can be challenging. Another approach is to train a deep neural network to describe and correct the geometric distortion. However, training a deep neural network that can handle a wide range of deformations while maintaining control over the geometric properties of the deformation is a challenging task. This difficulty often leads to inaccurate estimations of the required geometric distortion necessary to correct an image, making it crucial to develop a network that can effectively learn spatial deformation with controlled local geometric distortions.

\begin{figure}
    \centering
    \includegraphics[width=0.8\textwidth]{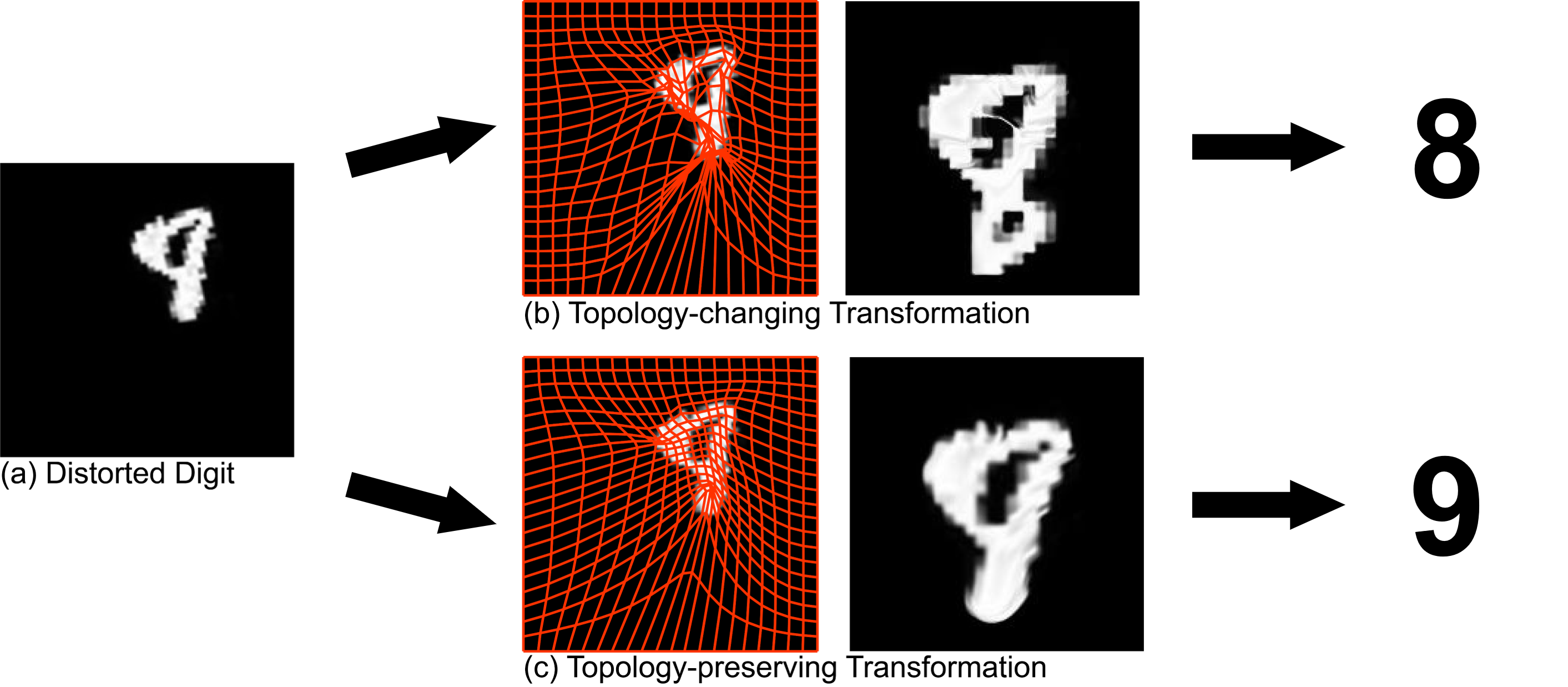}
    \caption{The significance of bijectivity. (a) A degraded image of the digit 9. (b) The degraded image undergoes a non-bijective deformation resulting in a topological change, transforming the digit 9 into the digit 8. (c) The degraded image undergoes a bijective deformation. The distorted digit 9 is transformed into a non-distorted digit 9.}
    \label{fig:diagnosour}
\end{figure}

To address the problem of imaging tasks for geometrically distorted images, we propose the deformation-invariant neural network (DINN), a framework that integrates the quasiconformal transformer network (QCTN) into existing deep neural networks. The QCTN is a deep neural network that outputs a quasiconformal map, which can transform a geometrically distorted image into an improved version that is closer to the distribution of natural or good images. The QCTN achieves this by first outputting a Beltrami coefficient, which measures the quasiconformality of the associated deformation map. By controlling the Beltrami coefficient, the local geometric distortion under the quasiconformal mapping can be controlled. A key feature of the QCTN is its ability to generate a bijective deformation map. The bijectivity holds great importance as it ensures the preservation of the essential characteristics of the original image. Figure \ref{fig:diagnosour} provides an illustration of the importance of bijectivity. In Figure \ref{fig:diagnosour}(a), an image representing a degraded digit 9 is shown. Our goal is to transform the degraded image into an undistorted version and position the digit at the center. Figure \ref{fig:diagnosour}(b) displays the transformed image using a non-bijective deformation map. It is important to note that the digit 9 undergoes a topological change and is transformed into the digit 8. Consequently, the transformed image is incorrectly recognized as a digit 8 by a classification network. On the other hand, Figure \ref{fig:diagnosour}(c) presents the transformed image using a bijective deformation map. The digit 9 is transformed into an undistorted digit 9. As a result, the transformed image can be accurately identified as a digit 9 by a classification network. This example highlights the critical role of bijectivity in preserving the essential characteristics of the original image. Additionally, from a mathematical perspective, this regularization of the output space helps mitigate the risk of overfitting by limiting the range of possible transformations. Consequently, it enhances the robustness and generalization capabilities of our network.

Utilizing our framework, we have devised an image classification network that excels in accurately classifying distorted images. Our proposed framework has also been applied to the restoration of geometrically distorted images, including images distorted by atmospheric turbulence and water turbulence. Our proposed framework has outperformed existing GAN-based restoration methods, demonstrating its effectiveness. Additionally, we have applied our proposed framework to the 1-1 verification of human face images under strong air turbulence and achieved good performance, further demonstrating the efficacy of our approach. The proposed DINN framework is believed to be an effective model to enhance the performance of deep neural networks in imaging tasks and enable robust and accurate image analysis in various applications.

In summary, the main contributions of this paper are listed below.

\han{
\begin{itemize}
    \item We introduce the Deformation-Invariant Neural Network (DINN) framework to address imaging tasks where geometrically distorted images are involved. As the core component is portable, large pre-trained networks can seamlessly handle heavily distorted images without requiring additional fine-tuning.
    
    \item Based on quasiconformal theories, the QCTN component in DINN generates a bijective deformation map, preserving the salient features of the original image. This property leads to more accurate imaging results, ensuring that the restored images maintain their essential characteristics.
    
    \item We utilize the DINN framework to design deep neural networks for tackling three imaging tasks. (1) image classification; (2) image restoration in the presence of atmospheric or underwater turbulence; (3) 1-1 facial verification tasks involving facial images corrupted by atmospheric turbulence.
\end{itemize}
}
\section{Related Work} In this section, we present a comprehensive overview of the relevant existing literature that closely relates to our work. 

\subsection{Computational quasiconformal geometry}
Computational quasiconformal geometry has found extensive application in diverse imaging tasks, yielding successful results. Computational quasiconformal geometry has provided a mathematical tool to study and control the geometric distortions under a mapping. In particular, conformal mappings belong to the class of quasiconformal mappings. Conformal mappings have garnered widespread usage in geometry processing, finding application in a multitude of tasks, including but not limited to texture mapping and surface parameterizations \cite{levy2002least,gu2004genus,gu2003global}. To quantitatively assess local geometric distortions within a mapping, the associated Beltrami coefficient is commonly employed. By manipulating the Beltrami coefficients, effective control over the geometric properties of the mapping can be achieved. Consequently, various surface parameterization methods minimizing conformality distortion have been proposed, leveraging the Beltrami coefficient~\cite{choi2016spherical,choi2020free}. Besides, quasiconformal mappings have found applications in computational fabrication~\cite{Soliman:2018:OCS,Crane:2013:RFC,panetta2019x}. Moreover, numerous quasiconformal imaging models have been introduced in recent years to address diverse imaging tasks, such as image registration ~\cite{lam2014landmark,lui2014teichmuller}, surface matching~\cite{choi2015fast} and shape prior image segmentation~\cite{zhang2021topology,siu2020image}. 

\subsection{Deformable Convolution}

Deformable convolution has emerged as a promising solution to overcome the limitations of traditional convolution operations in Convolutional Neural Networks (CNNs). One notable approach is the Active Convolution (AC) proposed by Y. Jeon et al. \cite{jeon2017active}. AC integrates a trainable attention mechanism into the convolution operation, enabling adaptive feature selection for different input instances. Another related technique is the Spatial Transformer Network (STN) \cite{jaderberg2015spatial}, which introduces a learnable transformation module capable of warping the input feature map based on learnable parameters. By incorporating an explicit spatial transformation module, the STN allows the network to learn spatial transformations that align the input with the task at hand, resulting in improved performance for tasks like digit recognition and image classification. Building upon the concept of the STN, D. Dai et al. \cite{dai2017deformable} proposed the Deformable Convolution Network (DCN), which extends the idea of spatial transformation to the convolution operation itself. DCN introduces learnable offsets for each position in the convolutional kernel, enabling dynamic adjustment of the sampling locations for each input instance. This leads to enhanced performance in tasks such as object detection and semantic segmentation. However, the original DCN has limitations in handling large deformations and maintaining invariance to occlusion. To address these limitations, researchers have introduced variations such as the Deformable Convolution v2 (DCNv2) \cite{zhu2019deformable}, which incorporates additional deformable offsets for intermediate feature maps, and the Deformable RoI Pooling (DRoIPool) \cite{dai2017deformable}, which extends the DCN to region-based object detection tasks. Furthermore, W. Luo et al. \cite{luo2016understanding} discovered that the contribution of each pixel is not equal to the final results in DCN, highlighting the need for further improvements in the deformable convolution operation to address its limitations and maximize its performance.

\subsection{Image Restoration}
GAN models have shown success in image restoration by training a generator to restore degraded images to their original, high-quality versions \cite{kupyn2018deblurgan, kupyn2019deblurgan, zhu2017unpaired, isola2017image}. The generator learns to produce visually pleasing and realistic images by taking a degraded image as input and generating an output that closely resembles the original image. In the specific area of image deturbulence, which focuses on restoring images distorted by atmospheric or water turbulence, Lau et al. \cite{lau2019restoration} propose a method that utilizes robust principal component analysis (RPCA) and quasiconformal maps. Their approach aims to restore atmospheric turbulence-distorted images. Another relevant work by Thepa et al. \cite{thapa2020dynamic} presents a deep neural network-based approach for reconstructing dynamic fluid surfaces from various types of images, including monocular and stereo images.  Anantrasirichai \cite{anantrasirichai2023atmospheric} designed a restoration network using complex-valued convolutions. Lau et al. \cite{lau2020atfacegan} propose a novel GAN-based approach specifically designed for restoring and recognizing facial images distorted by atmospheric turbulence. Their method can restore high-quality facial images from distorted inputs and recognize faces under challenging conditions. Li et al. \cite{li2018learning} employ a GAN-based model to remove distortions caused by refractive interfaces, such as water surfaces, using only a single image as input. \cite{Jiang_2023_CVPR} used general implicit neural representation for atmospheric and water turbulence removal in an unsupervised approach. Furthermore, Rai et al. \cite{rai2022removing} introduce a channel attention mechanism, adapted from \cite{hu2018squeeze}, into the generator network of their proposed GAN-based model. This mechanism helps the network focus on more relevant features during the restoration process.
\section{Mathematical Formulation}
In this section, we present the mathematical formulation of our proposed framework, which leverages the principles of quasiconformal geometry. We also provide an overview of the fundamental mathematical concepts related to quasiconformal theories.

\subsection{Problem formulation}
In this subsection, we provide the mathematical formulation of our problem. Given a distorted image $\tilde{I}:\Omega\to \mathbb{R}$ or $\mathbb{R}^3$, $\tilde{I}$ can be expressed as:

\begin{equation}
\tilde{I} = I \circ \tilde{f} + \epsilon,
\end{equation}

\noindent where $I$ is the original clean image without any refractive distortion, and $\epsilon$ represents the additive noise. Here, $\tilde{f}$ represents the geometric distortion imposed on $I$.

For most imaging tasks, the input image is assumed to be clean or undistorted. The clean image is fed into a suitable algorithm or deep neural network, which produces the desired imaging results. For example, in a classification task, we use a classifier network $\mathcal{N}$ that takes an image as input and outputs its predicted category. Usually, $\mathcal{N}$ is trained on a dataset of normal, undistorted images with a distribution denoted as $P_{\text{clear}}$. When the distorted image $\tilde{I}$ is used as input to $\mathcal{N}$, $\mathcal{N}(\tilde{I})$ is likely to produce an incorrect prediction since $\tilde{I}$ significantly deviates from $P_{\text{clear}}$. One possible solution is to adjust the parameters in the deep neural network using a training dataset of distorted images. However, in most cases, the network $\mathcal{N}$ is so large that it becomes expensive to include these distorted images for fine-tuning. Additionally, the introduction of extra variance in the data distribution caused by the distorted images may potentially degrade the performance of the tuned neural network.

To address this problem, our strategy is to learn a deformation map $f:\Omega \to \Omega$ to deform $\tilde{I}$ such that the deformed image $I' = \tilde{I}\circ f$ is closer to the distribution $P_{\text{clear}}$. In other words, $I' \sim P_{\text{clear}}$. Essentially, $f \approx \tilde{f}^{-1}$. To solve this problem, we propose the Quasiconformal Transformer Network (QCTN) to learn the deformation $f$. The QCTN is lightweight, making it more cost-effective than retraining $\mathcal{N}$ with a large dataset of distorted images.

\subsection{Quasi Conformal Theory}
As discussed in the previous subsection, our strategy for handling distorted images is to train the Quasiconformal Transformer Network (QCTN). Specifically, the QCTN learns a homeomorphic mapping $f:\Omega\to \Omega$ to remove the geometric distortion from a distorted image $\tilde{I}$. Each homeomorphic deformation $f:\Omega\to \Omega$ is associated with a geometric quantity known as the \textit{Beltrami coefficient}, defined as:

\begin{equation}\label{eq:beleq}
\mu=\frac{\partial f}{\partial \bar{z}} / \frac{\partial f}{\partial z},
\end{equation}

\noindent where $z=x+iy \in \mathbb{C}$ is a complex-valued point in the image domain, equivalent to $(x,y) \in \mathbb{R}^2$. $\frac{\partial}{\partial z} = \frac{1}{2}\left(\frac{\partial}{\partial x} -i\frac{\partial}{\partial y}\right)$ and $\frac{\partial}{\partial \bar{z}} = \frac{1}{2}\left(\frac{\partial}{\partial x} +i\frac{\partial}{\partial y}\right)$. Here, $i$ represents the imaginary unit.

The Beltrami coefficient $\mu:\Omega \to \mathbb{C}$ is a complex-valued function defined on the image domain $\Omega$. It quantifies the local geometric distortion under the mapping $f$. In particular, if $\mu(z) = 0$, then $\frac{\partial f}{\partial \bar{z}}(z) =0$, indicating that $f$ is conformal at $z$. According to quasiconformal theories, there exists a one-to-one correspondence between the space of homeomorphic mappings and the space of Beltrami coefficients. Given a homeomorphic deformation, its associated Beltrami coefficient $\mu$ can be obtained using Equation \ref{eq:beleq}. Conversely, given a Beltrami coefficient $\mu$ with $||\mu||_{\infty} <1$, the associated homeomorphic mapping $f$ can be reconstructed by solving Beltrami's equation. In particular, the condition $||\mu||_{\infty} <1$ ensures that the reconstructed mapping $f$ is bijective. In our framework, we aim to generate a bijective deformation $f$ to remove the geometric distortion while preserving the essential characteristics of the original image. This can be achieved by controlling the Beltrami coefficient, which measures the local geometric distortion of the associated mapping.

In our proposed model, the QCTN takes the input image $\tilde{I}$ and outputs the Beltrami coefficient $\mu$ associated with the desired deformation map $f$. To ensure the bijectivity of $f$, $\mu$ is constrained to satisfy $||\mu||_{\infty} <1$ by applying an activation function. The resulting $\mu$ is then fed into another network that outputs the mapping $f$ corresponding to $\mu$. The mapping $f$ is then used to remove the geometric distortion from $\tilde{I}$ for further processing.

\section{Deformation-invariant Neural Network (DINN)}

\begin{figure*}[t]
    \centering
    \includegraphics[width=\textwidth]{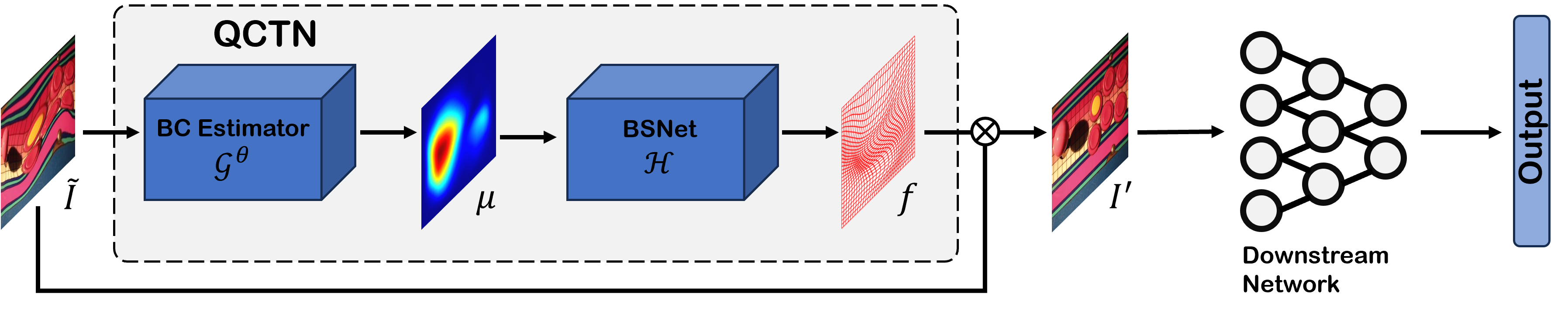}
    \caption{The overall framework of the proposed Deformation-invariant Neural Network (DINN). This framework consists of three principal modules: the Beltrami Coefficient (BC) estimator, the Beltrami Solver network (BSNet), and the network dedicated to a specific downstream imaging task.}
    \label{fig:modelpipeline}
\end{figure*}
In this section, we provide a comprehensive explanation of our general framework, known as the Deformation-invariant Neural Network (DINN), designed to address imaging tasks on distorted images. Each component of DINN will be discussed in subsequent subsections.

\subsection{Overall framework}
In this subsection, we present an overview of the Deformation-Invariant Neural Network (DINN) framework, illustrated in Figure \ref{fig:modelpipeline}. DINN begins by taking a distorted image as input, which is then processed by the Quasiconformal Transformer Network (QCTN). The QCTN, a lightweight neural network, generates a deformation map aimed at rectifying the geometric distortion present in the input image(s). Comprising two key components, the QCTN consists of (1) the Beltrami coefficient estimator and (2) the Beltrami Solver Network (BSNet). The Beltrami coefficient estimator computes the Beltrami coefficient $\mu$ from the distorted image input $\Tilde{I}$, while the BSNet utilizes the estimated coefficient $\mu$ to solve for its corresponding deformation mapping $f$. This mapping is applied to the distorted image $\Tilde{I}$, resulting in a less distorted version denoted as $I' = \Tilde{I} \circ f$, which aligns more closely with the distribution of clean, non-distorted images. The transformed image $I'$ is then passed to a downstream network tailored to the specific imaging task at hand. The subsequent subsection offers a detailed explanation of the Beltrami coefficient estimator, BSNet, and the loss function utilized for network training.

\subsection{Quasiconformal Transformer Network (QCTN)}
A crucial component in DINN is the incorporation of the Quasiconformal Transformer Network (QCTN). The distorted image is fed into the QCTN, which outputs a deformation map. This deformation map is then used to spatially transform the distorted image into one that is less distorted for further processing. In existing works \cite{li2018learning,rai2022removing}, the estimation of the restoration mapping is often performed without regularization, which can lead to overfitting. In our work, we introduce the concept of quasiconformality into the network to gain better control over the geometric properties of the deformation map. By incorporating regularization based on quasiconformality, we can ensure that the output space consists of quasiconformal mappings. As a result, the risk of overfitting can be effectively reduced, thereby significantly enhancing the robustness of our network. Specifically, our QCTN consists of two components: (1) the Beltrami coefficient estimator and (2) the Beltrami Solver network (BSNet).

The Beltrami coefficient estimator aims to estimate the Beltrami coefficient associated with the desired deformation map. The Beltrami coefficient measures the geometric distortion caused by the deformation map, allowing for easy control of the geometric properties of the deformation map. The BSNet then outputs the corresponding deformation map associated with the estimated Beltrami coefficient. In the following subsections, we will provide a detailed description of each component.

\bigskip

\subsubsection{Beltrami coefficient estimator} 
\begin{figure}[t]
    \centering
    \includegraphics[width=0.85\textwidth]{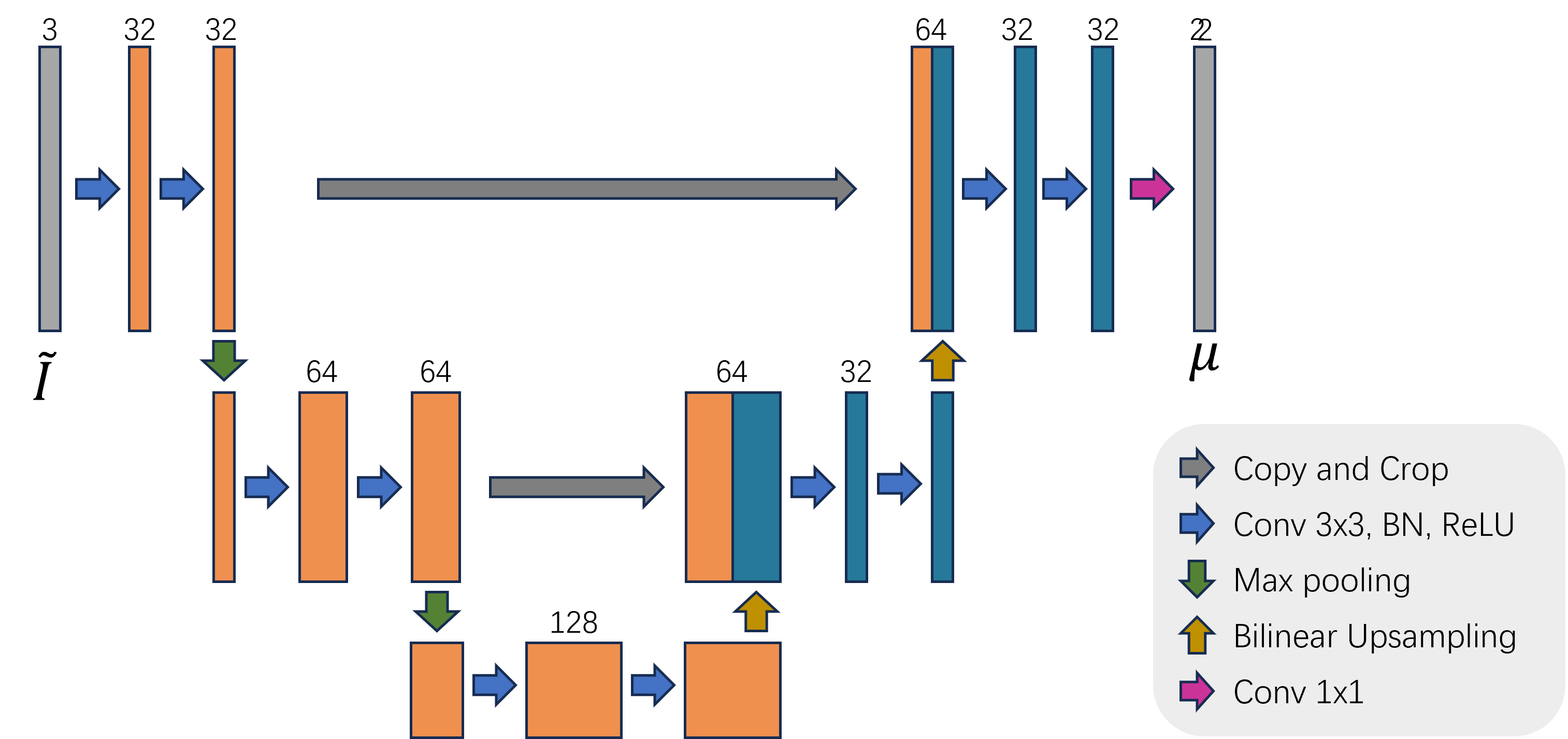}
    \caption{The architecture of the Beltrami coefficient (BC) estimator.}
    \label{fig:BCE}
\end{figure}
The main feature of the QCTN is the utilization of the Beltrami coefficient (BC) to represent the deformation map, as opposed to the conventional approach of using the vector field. The BC, denoted as $\mu$, quantifies the local geometric distortion caused by the deformation map. Specifically, if the norm of $\mu$, denoted as $|\mu|$, is close to zero, it indicates that the geometric distortion under the associated deformation map is minimal. Therefore, a loss function can be designed to minimize $|\mu|$ in certain regions, thereby reducing the local geometric distortion.

Moreover, for a bijective deformation, the BC satisfies the condition $||\mu||_{\infty} < 1$. In this work, generating a bijective deformation map is crucial for mitigating the geometric distortion in the distorted image. To ensure that the mapping outputted by QCTN is bijective, we employ the activation function \eqref{eq:BCactivation}. This activation function is applied to ensure that the Beltrami coefficient, which is outputted by the Beltrami Estimator Network, has a supreme norm strictly less than 1. By enforcing this condition, we guarantee the bijectivity of the resulting deformation map. In the discrete case, the image domain is triangulated, and each deformation mapping is treated as a piecewise linear function over each triangular face. The first derivatives of a piecewise linear function are constant on each triangular face. Consequently, the Beltrami coefficient can be regarded as a complex-valued function defined over each triangular face.

As shown in Figure \ref{fig:BCE}, the Beltrami coefficient estimator is an encode-decoder network $\mathcal{G}^{\theta}$ with trainable parameters $\theta$, which takes distorted image $\Tilde{I}$ as input and outputs a BC $\mu\in \mathbb{C}^m$ associated with a deformation map $f$ for restoring the distorted image. Here, $m$ is the number of triangular faces in the discretization of the image domain. The Beltrami coefficient $\mu=\rho+i\tau$ defined in the image domain can be treated as two-channel images, where one channel represents the real part $\rho=Re(\mu)$, and the other one represents the imaginary part $\tau=Im(\mu)$. The corresponding deformation map $f$ associated with the output BC $\mu$ can be obtained by the BSNet $\mathcal{H}$, which will be described in the next subsection. This mapping $f = \mathcal{H}(\mu)$ is then used to transform the distorted image $\Tilde{I}$ into a distortion-free one $I'=\Tilde{I}\circ f$. In order to obtain a bijective deformation map, the BC $\mu$ outputted by $\mathcal{G}^{\theta}$ should satisfy the condition that $||\mu||_{\infty}<1$. For this purpose, in the last layer of the Beltrami coefficient estimator, we apply the following activation function:
\begin{equation}
    \mathcal{A}(\mu_j) = \frac{e^{|\mu_j|} - e^{-|\mu_j|}}{e^{|\mu_j|} + e^{-|\mu_j|}}\textbf{arg}(\mu_j),
    \label{eq:BCactivation}
\end{equation}
where $\mu_j$ is the $j$-th entry of $\mu \in \mathcal{C}^m$ and $\textbf{arg}(\mu_j)$ is the argument of the complex number $\mu_j$.

The activation function $\mathcal{A}$ guarantees that the network produces output values $\mu$ with a supremum norm that is strictly less than 1. As a result, the associated deformation map becomes bijective. This activation function plays a crucial role to ensure a bijective deformation map in our DINN framework. The parameters of $\mathcal{G}^{\theta}$ are optimized by backward propagation to minimize suitable loss functions, which will be described in subsection \ref{training}.

\bigskip

\subsubsection{Beltrami Solver Network (BSNet)} 
\begin{figure}[t]
    \centering
    \includegraphics[width=1\textwidth]{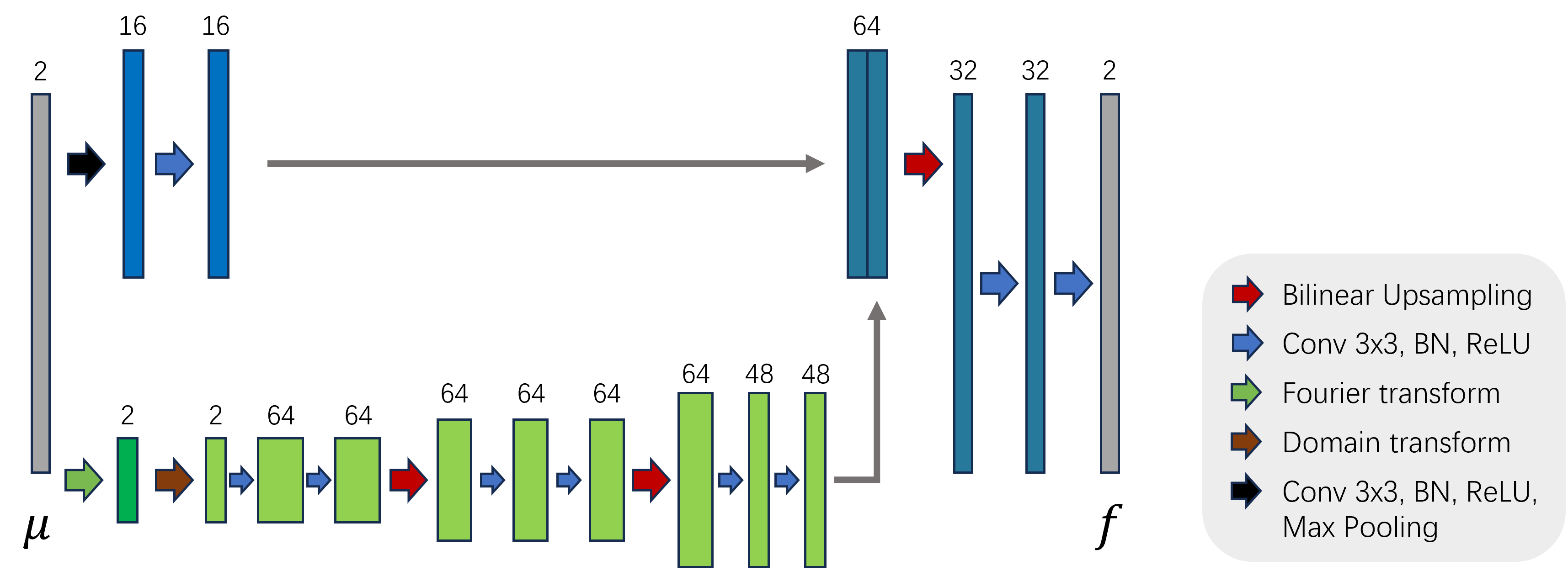}
    \caption{The architecture of the Beltrami Solver Network}
    \label{fig:BSnet}
\end{figure}
Another component in the QCTN is the pretrained Beltrami Solver Network (BSNet). The BSNet, denoted by $\mathcal{H}$, takes the BC $\mu\in \mathbb{C}^m$ as input and outputs its corresponding deformation map $f = \mathcal{H}(\mu)$. Mathematically, the BSNet solves Beltrami's equation:
\begin{equation}
    \frac{\partial f}{\partial \bar{z}} = \mu \frac{\partial f}{\partial z}.
\end{equation}

Beltrami's equation has a variational formulation and can be converted into a system of elliptic partial differential equations:
\begin{equation}\label{ellipticPDE}
    \nabla \cdot \left(A \begin{pmatrix}
        u_x\\
        u_y
    \end{pmatrix}\right) = 0;\ \ \nabla \cdot \left(A \begin{pmatrix}
        v_x\\
        v_y
    \end{pmatrix}\right) = 0,
\end{equation}
where $A = \begin{pmatrix}
    \alpha_1 & \alpha_2\\
    \alpha_2 & \alpha_3
\end{pmatrix}$, $\alpha_1 = \frac{(\rho-1)^2+ \tau^2}{1-\rho^2 - \tau^2}$, $\alpha_2 = \frac{-2\tau}{1-\rho^2 - \tau^2}$, $\alpha_3 = \frac{(\rho+1)^2+ \tau^2}{1-\rho^2 - \tau^2}$, and $\mu = \rho + i\tau$, where $\rho$ and $\tau$ are the real and imaginary parts of the Beltrami coefficient $\mu$. Note that $\mu$ is defined at every $z=x+iy$, whose associated coordinate in the image domain is $(x,y)$. In the discrete case, the image domain $\Omega$ is discretized by a triangulation mesh, and $f$ is piecewise linear on each triangular face. Suppose $f = {\bf u} + i{\bf v}$, where ${\bf u}$ and ${\bf v}$ are the coordinate functions of $f$ defined on every vertex. Then, the system of elliptic PDEs can be discretized into two sparse linear systems: $C_1{\bf u}={\bf 0}$ and $C_1{\bf v}={\bf 0}$. These can be used to define a loss function to train the BSNet:
\begin{equation}\label{BSNetLoss}
    \mathcal{L}_{BSNet} = ||C_1{\bf u}||_1 + ||C_2{\bf v}||_1.
\end{equation}

The architecture of the BSNet is shown in \ref{fig:BSnet}. The network consists of a short path (upper network) and a long path (bottom network). The goal is to design a smaller network with fewer parameters for efficient training. To achieve this goal, we consider the Fourier transform of $\mu$. This is inspired by the property that the low-frequency component of $\mu$ can effectively capture the overall pattern of the corresponding deformation map $f$ \cite{lui2013texture}. In the long path, we first perform the discrete Fourier transform on $\mu$, and then apply truncation to the Fourier coefficient matrix to keep a few coefficients associated with the low-frequency component. The truncated Fourier coefficient matrix contains the major information of the corresponding mapping. After that, the truncated Fourier coefficient matrix is fed into the Domain Transform Layer (DTL), which imitates the process of transforming features from the frequency domain to the spatial domain. The network is then extended with multiple convolutional layers, each followed by an activation function. The truncation of the Fourier coefficient matrix greatly reduces the number of variables and parameters in the network. However, some subtle information may be lost due to the truncation of high-frequency components of the Fourier coefficients. Specifically, some local deformation patterns may be lost. To address this, we add a short path, which consists of a few layers of convolution and downsampling. The output is concatenated with the output from the long path. This short path is shallow and has a minimal training burden on the overall network. More details about the BSNet can be found in \cite{chen2021deep}.

\subsection{Training process of DINN}\label{training}
In the DINN framework, our main task is to train the Beltrami coefficient estimator and the BSNet. The Beltrami coefficient estimator is lightweight, making the training process cost-effective. The BSNet can either be pretrained or trained simultaneously, depending on the application. The downstream network for performing a specific imaging task is pretrained using the clean, undistorted training dataset. For the ease of our discussion, denote the Beltrami coefficient estimator, BSNet, and the downstream network for the imaging task as $\mathcal{N}_\theta$, $\mathcal{H}_\phi$, and $\mathcal{T}_\varphi$, respectively.

To train the DINN framework, we optimize the loss function with suitable loss terms associated with each component of the DINN. The loss function $\mathcal{L}$ is in the form:
\begin{equation}
   \mathcal{L}(\theta,\phi) = \alpha \mathcal{L}_{est} + \beta \mathcal{L}_{BSNet} + \gamma \mathcal{L}_{task}.
\end{equation}

$\mathcal{L}_{est}$ guides the training of the Beltrami coefficient estimator by enforcing it to output a Beltrami coefficient associated with a suitable deformation map, which aligns the deformed image with its ground truth. For instance, suppose the distorted images and their corresponding ground truth images are available in the training data. Let $\Tilde{I}$ be the distorted image and $I$ be its distortion-free ground truth. Then, $\mathcal{L}_{est}$ can be designed to measure the mean square error between the deformed image and the ground truth image:
\begin{equation}
    \mathcal{L}_{est} = ||\Tilde{I}\circ \mathcal{H}_{\phi}\circ \mathcal{N}_\theta(\Tilde{I}) - I||_2.
\end{equation}

In some applications where the ground truth deformation map $f_{\Tilde{I}}$ to restore the distorted image $\Tilde{I}$ is known, $\mathcal{L}_{est}$ can be designed as:
\begin{equation}
    \mathcal{L}_{est} = ||\mathcal{H}_{\phi}\circ \mathcal{N}_\theta(\Tilde{I}) - f_{\Tilde{I}}||_2.
\end{equation}

Additionally, $\mathcal{L}_{BSNet}$ guides the training of the parameters $\phi$ of the BSNet, ensuring that $\mathcal{H}$ solves Beltrami's equation. This network can also be pretrained and frozen after the pertaining. Then, the corresponding weighting parameter can be set as $\beta = 0$. In practice, we choose to pre-train BSNet separately, as it is specifically designed to solve the Beltrami equation given a Beltrami coefficient. Furthermore, jointly training all networks can result in a more complex optimization problem. By pre-training BSNet individually, we simplify the training process while enhancing its ability to accurately solve the equation. Furthermore, pre-training BSNet facilitates faster convergence overall and yields a better performance.

Finally, $\mathcal{L}_{task}$ is the loss function used to train the downstream network for the specific imaging task. It is included in the loss $\mathcal{L}$ to guide the output deformation map $f = \mathcal{H}_{\phi}\circ \mathcal{N}_\theta(\Tilde{I})$ such that the deformed image $I' = \Tilde{I}\circ f$ lies within the distribution of clean, undistorted images. By adding $\mathcal{L}_{task}$ to the loss $\mathcal{L}$, we aim to find $f$ such that the deformed image $I'$ produced by $f$ gives an accurate imaging result when fed into the pretrained downstream network $\mathcal{T}_\varphi$. Note that $\mathcal{T}_\varphi$ is pretrained on the training dataset of undistorted images. Thus, minimizing $\mathcal{L}_{task}$ encourages $f$ to deform $\Tilde{I}$ to one that aligns with the training dataset of undistorted images. For example, for image classification, $\mathcal{L}_{task}$ can be chosen as the cross-entropy of the probability vectors. If $I'$ remains distorted, $\mathcal{T}_\varphi$ is likely to produce an incorrect probability vector, resulting in a large cross-entropy with the correct probability vector (the label). By minimizing the cross-entropy, we guide the Beltrami coefficient estimator $\mathcal{N}_{\theta}$ to output a Beltrami coefficient associated with a deformation map $f$ that restores the deformed image $I'$ to an image that aligns with the training dataset of undistorted images of its corresponding class. The cross-entropy between $\mathcal{T}_\varphi(I')$ and the ground truth probability vector will be small.

In the following section, we will demonstrate the application of the DINN framework to real imaging tasks in order to illustrate the concept of the framework.

\section{Applications of DINN}
In this section, we describe how we can apply the DINN framework to three real applications, namely, (1) image classification; (2) image restortation and (3) 1-1 facial verification.

\begin{figure}[t]
    \centering
    \includegraphics[width=\textwidth]{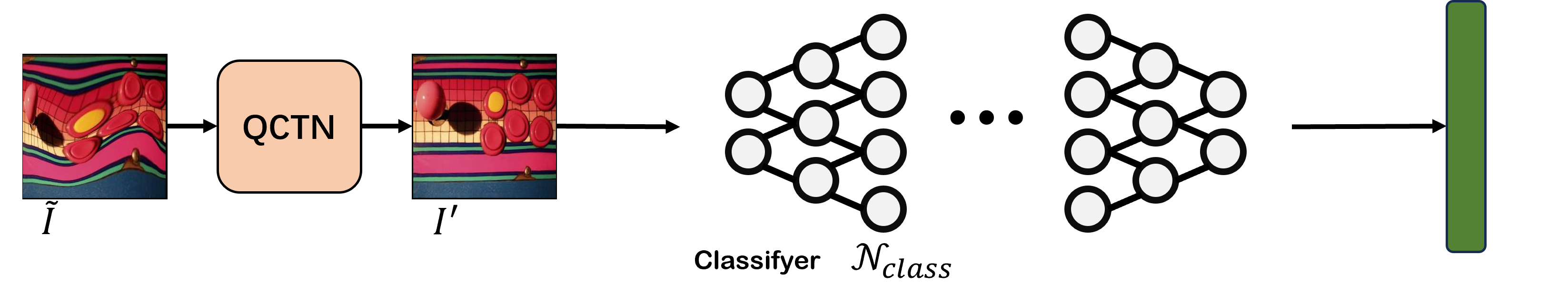}
    \caption{DINN based image classification network.}
    \label{fig:task1}
\end{figure}
\subsection{Classification of distorted images}\label{classification} The DINN framework can be applied to perform image classification on distorted images. In certain real-world scenarios, images can undergo geometric distortions. For example, images captured using a long-range camera may experience geometric distortions caused by atmospheric turbulence. When dealing with distorted images, the predictions made by a classification network $\mathcal{T}_{c}$ can be inaccurate due to the mismatch between the distorted image $\Tilde{I}$ and the training dataset, which consists of clean and undistorted images. To tackle this issue, we utilize the DINN framework to develop a deep neural network specifically designed for classifying distorted images. The main idea is to incorporate the QCTN component before the downstream classification network. The overall network architecture is illustrated in Figure \ref{fig:task1}. Initially, a distorted image $\Tilde{I}$ is inputted into the QCTN, which then produces a deformation map $f$. The deformed image $I'$ by $f$ is subsequently fed into the pre-trained downstream classification network, resulting in a probability vector.

Suppose we are given a training dataset of distorted images $\Tilde{I}$, whose classification labels are known. In certain situations, a training dataset may also be generated using physical experiments or mathematical simulations, resulting in paired images that comprise distorted images and their corresponding original counterparts. For example, one approach is to submerge a ground truth image in water and capture the image, accounting for water turbulence, to generate distorted images. This training dataset serves the purpose of guiding or initializing the parameters of the QCTN. To train the network, we optimize the parameters so that they minimize the following loss function:
\begin{equation}
    \mathcal{L}_c = \alpha \mathcal{L}_{est} + \beta \mathcal{L}_{BSNet} + \gamma \mathcal{L}_{ce},
    \label{eq:classification}
\end{equation}
\noindent where the task loss $\mathcal{L}_{task}= \mathcal{L}_{ce}$ is the cross entropy loss given by $\sum_{i=1}^n p_i \log q_i$, where ${\bf p} = (p_i)_{1\leq i\leq n}$ is the ground truth probability vector and ${\bf q} = (q_i)_{1\leq i\leq n}$ is the predicted probability obtained by the classification network $\mathcal{T}_{c}$. In case the original undistorted images corresponding to the distorted images in the training dataset are unavailable, we can set $\alpha =0$ and train the Beltrami coefficient estimator only by the task loss $\mathcal{L}_{ce}$ for the classification tasks. The BSNet can be pretrained, in which case $\beta$ is set to 0 and the BSNet is frozen. As mentioned in the previous subsection, $\mathcal{L}_{ce}$ guides the Beltrami coefficient estimator $\mathcal{N}_{\theta}$ to produce a Beltrami coefficient associated with a deformation map $f$. This deformation map is responsible for transforming the distorted image $I'$ into its undistorted version, aligning it with the distribution of clean, undistorted images.

\begin{figure}[t]
    \centering
    \includegraphics[width=\textwidth]{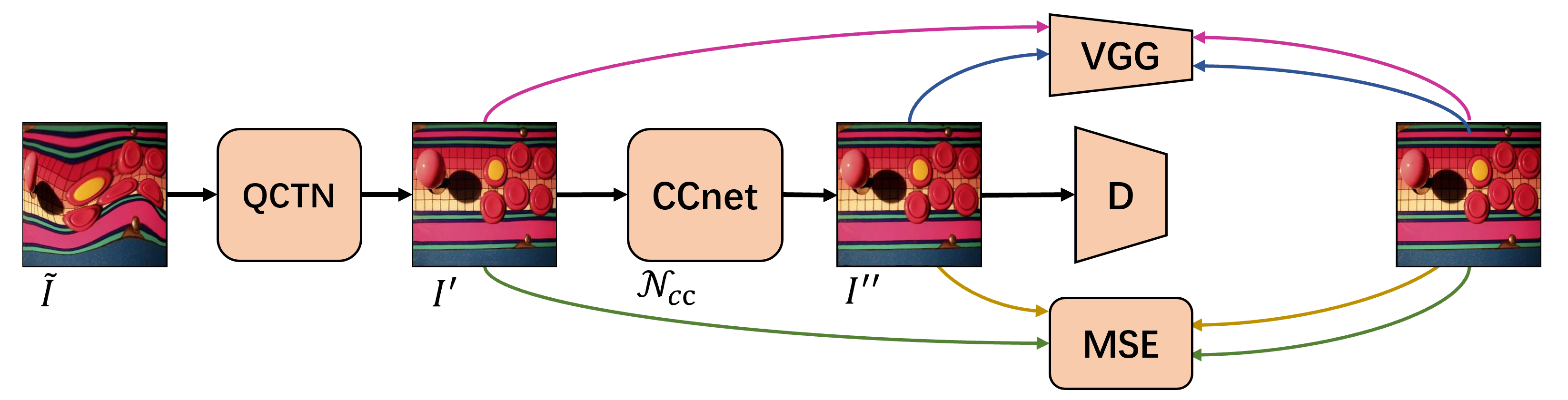}
    \caption{DINN based image restoration network.}
    \label{fig:task2}
\end{figure}
\subsection{Image restoration}\label{imagerestoration} The DINN can be applied to image restoration for turbulence-distorted images. These images can be affected by atmospheric turbulence or water turbulence. Our proposed model leverages a GAN-based architecture, as depicted in Figure \ref{fig:task2}. The deep neural network comprises the QCTN and an image deblurring module. The image deblurring module aims to restore a blurry image. Besides, turbulence-distorted images exhibit geometric distortions. The QCTN aims to eliminate the geometric distortions. Initially, the geometrically distorted image $\Tilde{I}$ is inputted into the QCTN, which generates an appropriate deformation map. The resulting deformed image, denoted as $I'$, is an improved version with the geometric distortions eliminated. However, it is possible for $I'$ to suffer from blurring caused by the spatial resampling process. To address this, $I'$ is further processed by a color correction network $\mathcal{N}^{\theta}_{cc}$, resulting in a color-corrected image referred to as $I''$. By combining the outputs of the QCTN and CCnet, we obtain the restored image $I''$, which serves as the generator within the GAN framework. Additionally, a discriminator $D$ is trained to evaluate the quality of the generated image and provide feedback to the generator. Throughout the training process, the generator aims to enhance the quality of the generated image, while the discriminator strives to distinguish between the restored and undistorted images.

In this work, our training dataset consists of pairs of geometrically distorted images and their corresponding undistorted originals. To train the network, we optimize the parameters so that they minimize the following loss function:
\begin{equation}
\begin{aligned}  
    \mathcal{L} = a_1 \mathcal{L}_{est} &+ a_2 \mathcal{L}_{MSE}(I,I^{''})+ a_3 \mathcal{L}_{vgg}(I,I^{'})  \\&+ a_4 \mathcal{L}_{vgg}(I,I^{''}) + a_5 \mathcal{L}_{adv}(I^{''}),
    \label{eq:restoration}
\end{aligned}
\end{equation}
\noindent Here, $\mathcal{L}_{MSE}(I,I^{''}) = ||I-I^{''}||_2^2$ is introduced in the loss function to guide the training of the CCnet, similar to the functionality of $\mathcal{L}_{est}$ for QCTN. $\mathcal{L}_{vgg}$ denotes the VGG loss, which measures the mean square error between the VGG features of two images:
\begin{equation}
\begin{aligned}  
    \mathcal{L}_{vgg}(I_1,I_2) = ||\Phi_{vgg}(I_1) - \Phi_{vgg}(I_2) ||^2,
\end{aligned}
\end{equation}
\noindent $\Phi_{vgg}(I_1)$ and $\Phi_{vgg}(I_2)$ denote the VGG features of $I_1$ and $I_2$ respectively. The VGGNet is pretrained on the ImageNet dataset for a classification task. The output of the last hidden layer is extracted and utilized as the VGG feature. Hence, the objective of the third and fourth terms is to encourage the VGG features of $I'$ and $I''$ to closely resemble the VGG feature of the ground truth image $I$. The last term $\mathcal{L}_{adv}(I,I^{''})$ is the adversarial loss defined as:
\begin{equation}
    \mathcal{L}_{adv}(I,I^{''}) = \log(D(I))+ \log (1 - D(I'')),
\end{equation}
\noindent where $D$ is the discriminator for determining whether an input image is real or generated, yielding output values between 0 and 1, respectively. By minimizing the adversarial loss, the objective is to guide the synthesized image $I''$ towards closely resembling a clean, undistorted image that the discriminator recognizes as real. In this task, the CCnet serves as a deblurring module and can be pretrained. The discriminator plays a crucial role in the min-max game of the GAN model and needs to participate in the training process. The optimization of the network parameters then follows an alternating approach, similar to the conventional GAN model. In this paper, we have set the weight parameters as $a_1=1.0$, $a_2=1$, $a_3=0.5$, $a_4=0.1$, and $a_5=0.1$ for our experiments.

\begin{figure}[t]
    \centering
    \includegraphics[width=\textwidth]{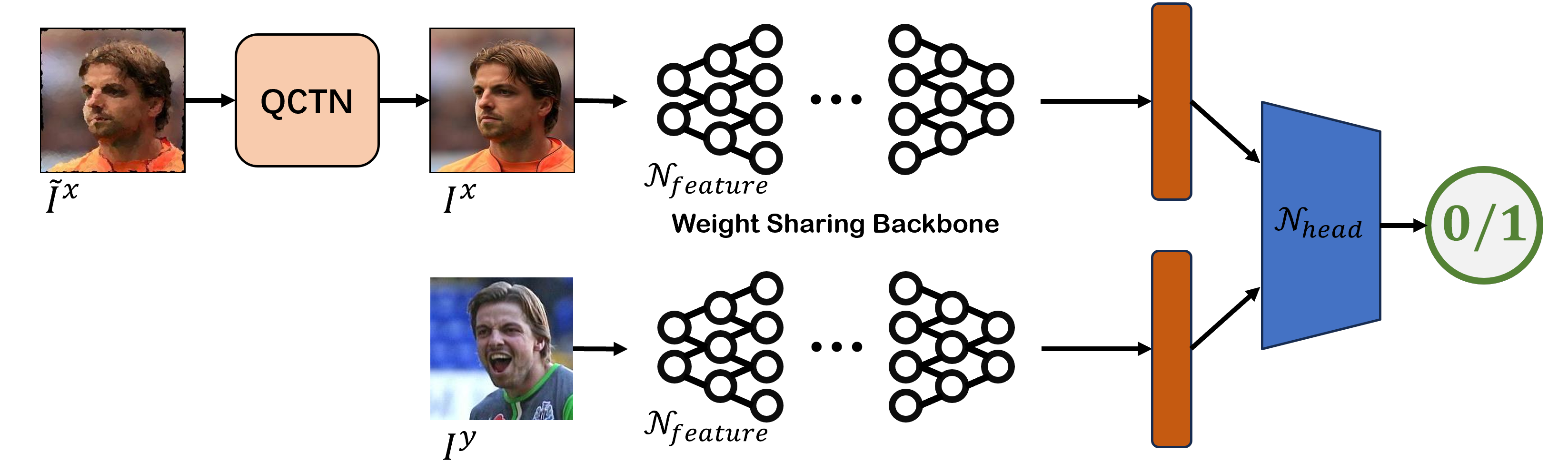}
    \caption{DINN based 1-1 facial verification network.}
    \label{fig:task3}
\end{figure}
\subsection{1-1 facial verification}\label{1-1facialverification} The DINN framework can also be applied to the problem of 1-1 facial verification. In this task, the objective is to determine whether a distorted facial image $\Tilde{I}^x$ belongs to the same person as another facial image $I^y$. In real-world scenarios, facial images captured by long-range cameras often exhibit geometric distortion caused by air turbulence, particularly when considering zoomed-in images. Our proposed model is illustrated in Figure \ref{fig:task3}. Initially, the distorted image $\Tilde{I}^x$ is inputted into the image restoration network, as described in the previous subsection, resulting in a geometrically restored and color-corrected image $I^x$. Subsequently, both the restored image $I^x$ and the reference image $I^y$ are fed into a feature extractor network, denoted as $\mathcal{N}_{feature}$, which utilizes the \textit{IR-50} architecture \cite{he2016deep}. The feature representations $\mathcal{N}_{feature}(I^x)$ and $\mathcal{N}_{feature}(I^y)$ are then passed to a similarity measure network, denoted as $\mathcal{N}_{head}$, which employs the ArcFace similarity comparison method \cite{deng2019arcface}. The final output of the model determines whether the two facial images belong to the same person, with a value of 1 indicating a match and 0 indicating a mismatch. During the training process, both $\mathcal{N}_{feature}$ and $\mathcal{N}_{head}$ are pre-trained. On the other hand, the image restoration network, which includes the QCTN, is trained according to the methodology described in the previous subsection. 

\section{Experimental Results}
In this section, we evaluate the efficacy of our proposed DINN framework through a series of experiments. Specifically, we assess the performance in image classification, image restoration, and 1-1 facial verification tasks involving distorted images using the DINN framework. We compare our results against those achieved by state-of-the-art methods. Additionally, we conduct self-ablation studies to explore the impact of various parameters and settings on the performance of the framework.

The experimental setting is described in detail below.

\medskip

\noindent\textbf{Computational Resources and Parameters}
Our models were trained using the RMSprop optimizer with a fixed learning rate of $0.00001$. The training underwent 100 epochs of optimization to achieve sufficient convergence. The batch size was set to 64, unless stated otherwise. The training process took place on a CentOS 8.1 central cluster computing node equipped with two Intel Xeon Gold 5220R 24-core CPUs and two NVIDIA V100 Tensor Core GPUs.

\medskip

\noindent\textbf{Training Details} For the classification task, the classifier network is pre-trained and kept fixed. Subsequently, the QCTN module is trained using the loss function defined in Equation \eqref{eq:classification}. In the image restoration and 1-1 facial verification tasks, the overall model consists of multiple components: the Beltrami coefficient Estimator, the Color Correction Network (CCnet), and the discriminator. The training process follows an alternating minimizing approach. 

\subsection{DINN for classification of distorted images}
\label{sec:MNIST}

In this subsection, we provide the experimental results of image classification for distorted images using the method introduced in subsection \ref{classification}. We evaluate the performance of the method on images distorted by different types of spatial deformations, specifically (1) affine transformations, (2) elastic transformations, and (3) a combination of affine and elastic transformations. The objective is to assess the capability of the proposed DINN framework in effectively handling various types of deformations.

\begin{table}[h!]
\caption{The classification accuracy of distorted images using different methods.}
\label{tb:CIFAR_CONV}
\label{tb:MNIST}
\label{tb:CIFAR_DEF}
\label{tb:FaMNIST}
\centering
\begin{tabular}{c|c|ccc}
\toprule
Deform Type             & Method    & Train ACC & Test ACC  & Invertible\\\hline
\multirow{3}{*}{Affine} & CNN       & 91.62     & 82.73     & $\setminus$\\ 
                        & STN    & 97.97     & 94.90     & \cmark\\ 
                        & DINN  & 97.45     & \textbf{96.32}     & \cmark\\\hline
                        
\multirow{3}{*}{Elastic}& CNN       & 95.43     & 78.47     & $\setminus$\\ 
                        & TPS-STN   & 99.37     & 81.94     & \xmark\\ 
                        & DINN  & 99.11     & \textbf{84.58}     & \cmark\\\hline
                        
                        & CNN       & 76.77     & 70.29     & $\setminus$\\ 
Affine \&	            & STN   & 81.65     & 77.21     & \cmark\\ 
Elastic                 & TPS-STN   & 86.13     & 80.63     & \xmark\\ 
                        & DINN  & 86.48     & \textbf{83.06}     & \cmark\\\hline
\bottomrule
\end{tabular}
\end{table}

\textbf{Affine Deformation} We evaluated the performance of our proposed method on the MNIST handwriting dataset distorted by affine transformations. To introduce these deformations, we applied rotation angles within the range of $[-\frac{\pi}{3},\frac{\pi}{3}]$ and scaling parameters within the range of $[0.2,0.6]$, while allowing translations within the image domain. The downstream image classification network utilized in our experiments is a convolutional neural network consisting of three convolutional layers and two fully connected layers. In our proposed model, the QCTN is appended before the downstream classification network to correct the geometric distortions. For comparison purposes, we also considered a related framework that employs the spatial transformer network (STN) \cite{jaderberg2015spatial}. In this alternative model, the STN is appended in front of the downstream classification network instead of the QCTN. 

During the classification stage, models equipped with transformer layers demonstrated better performance in terms of classification accuracy on the test dataset compared to the baseline convolutional neural network. The classification accuracy was notably low when the transformer layer was omitted. However, a significant improvement in classification accuracy was observed when the STN was added before the classification network. Furthermore, the incorporation of the QCTN resulted in even higher classification accuracy. A summary of the results can be found in Table~\ref{tb:MNIST}. The models were trained for 100 epochs to ensure convergence.

\begin{figure}[t]
    \centering
    \includegraphics[width =0.75\textwidth]{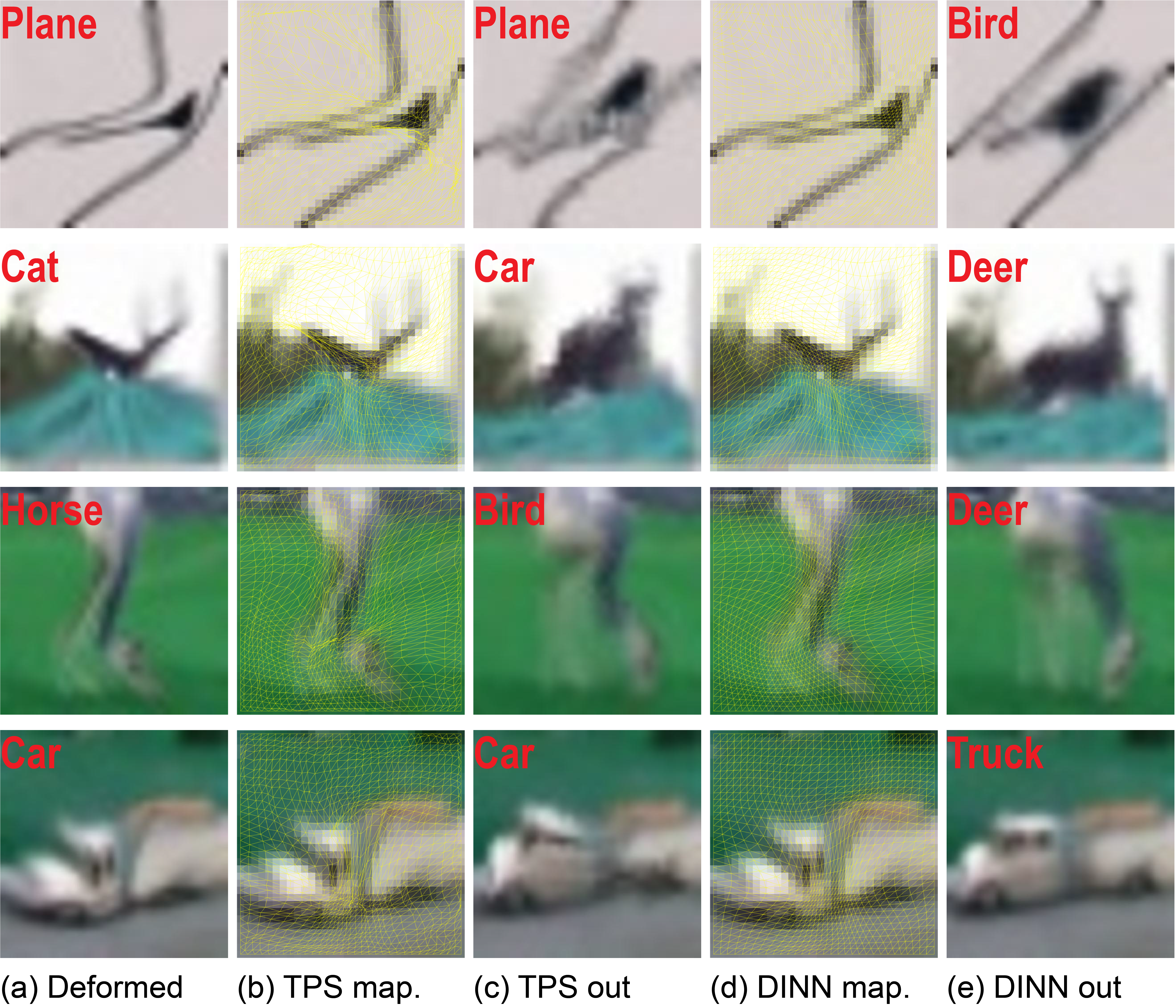}
    \caption{Images produced by the deformation maps from different transformer layers. (a) Distorted images. (b) Visualization of the mapping generated by TPS-STN on a deformed image. (c) Image recovered by TPS-STN. (d) Visualization of the mapping generated by QCTN on a deformed image. (e) Image recovered by QCTN. The class names in the top-right corner of images in columns (a), (c), and (e) indicate the predicted class by the baseline CNN, TPS-STN, and DINN, respectively. The correct labels from the top row to the last row are Bird, Deer, Deer and Truck.}
    \label{fig:CIFARnonrigid}
\end{figure}

\textbf{Elastic Deformation} We assessed the performance of our proposed method on images distorted by general elastic spatial deformations. Elastic deformations are commonly encountered in various scenarios, such as capturing images across uneven surfaces like glasses or water. To test the effectiveness of our method, we conducted experiments on the CIFAR10 dataset with large elastic deformations. Figure~\ref{fig:CIFARnonrigid} shows some examples of the deformed images. For this experiment, we employed the deep layer aggregation model (DLA) \cite{yu2018deep} as our downstream classification network. To ensure a fair comparison, we implemented a variant of the spatial transformer network (STN) with thin-plate spline transformation, called the {\it TPS-STN}. This variant outputs the mapped coordinates of the control points without constraint. The architecture for the TPS-STN that predicts the mapped coordinates is the same as our Beltrami coefficient estimator. The primary distinction between QCTN and TPS-STN is that QCTN generates a bijective folding-free deformation, whereas TPS-STN does not possess this property. The bijectivity of QCTN plays a crucial role in this imaging task.

Table~\ref{tb:CIFAR_DEF} presents the classification accuracy results of the downstream classification network, the classification network with TPS-STN, and the classification network with QCTN. Again, the classification models equipped with the transformer network exhibit significantly better accuracy. However, the use of QCTN yielded better results compared to TPS-STN. This is attributed to the non-bijective nature of the deformations produced by TPS-STN, which have impacted its performance.

Figure \ref{fig:CIFARnonrigid} displays several examples of deformed images generated by the deformation maps produced by the transformer layer. The aim is to ensure that the deformed images effectively alleviate the geometric distortions present in the input distorted images. The results demonstrate that the images restored by the DINN approach are notably more accurate and closely resemble an image from their respective classes. This improved restoration aids the classification network in better identifying the class to which each image belongs. The top left corner label for each image indicates the class recognized by the classification network. It is observed that both the downstream classification network and the network with the inclusion of STN provided incorrect classifications. However, with the integration of QCTN, the classifications were accurate.

\begin{figure*}[t]
    \centering
    \includegraphics[width =\textwidth]{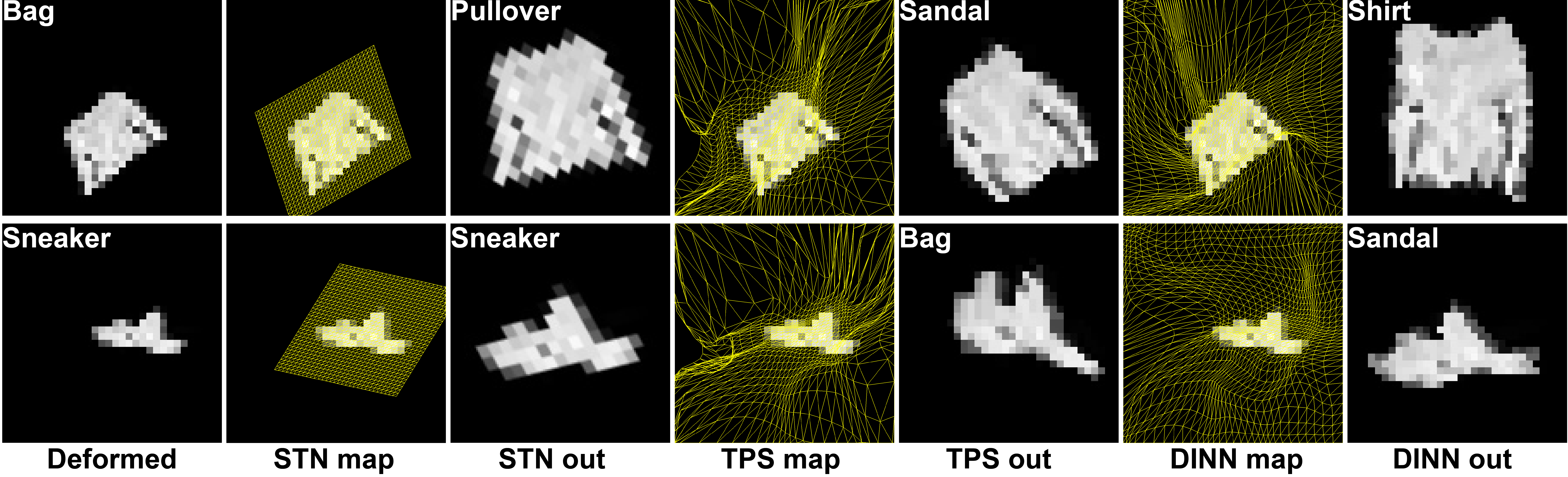}
    \caption{Images produced by the deformation maps from different transformer layers. (a) Distorted images from FashionMNIST. (b) Visualization of the mapping generated by STN on a deformed image. (c) Image recovered by STN. (d) Visualization of the mapping generated by TPS-STN on a deformed image. (e) Image recovered by TPS-STN. (f) Visualization of the mapping generated by QCTN on a deformed image. (g) Image recovered by QCTN. The class names in the top-right corner of images in columns (a), (c), (e) and (g) indicate the predicted class by the baseline CNN, STN, TPS-STN, and DINN, respectively. The correct labels from the top row to the last row are Shirt and Scandal.}
    \label{fig:FaMNISTnonrigid}
\end{figure*}
\textbf{Combined Deformation} 
We further evaluate the performance of our proposed model on images distorted by a combination of elastic and affine deformations, which generally involve large deformations. The experiment is conducted on the FashionMNIST dataset, utilizing the same downstream classification network as in the experiments on images distorted by affine transformations. We compare our method with the downstream classification network, the STN network, and the TPS-STN network. The classification accuracies are presented in Table~\ref{tb:FaMNIST}. Once again, the classification models equipped with the transformer network demonstrate significantly improved accuracy. However, the use of QCTN yields notably superior results compared to both STN and TPS-STN. Even for such large deformations, our proposed model successfully preserves the bijectivity of the deformation map, whereas TPS-STN fails to do so.

Figure \ref{fig:FaMNISTnonrigid} shows several examples of deformed images generated by the deformation maps produced by the transformer layer. Each image's top-left corner label indicates the class recognized by the classification network. The results demonstrate that the images restored by the DINN approach are notably more accurate and closely resemble an image from their respective classes, even with such large deformations. It is observed that both the downstream classification network, the network with the inclusion of STN and the network with the inclusion of TPS-STN, provided incorrect classifications. On the other hand, the classifications were accurate under the DINN framework.

\subsection{Restortion of Turbulence-distorted images}

\begin{figure}[t]
    \centering
    \includegraphics[width=.7\textwidth]{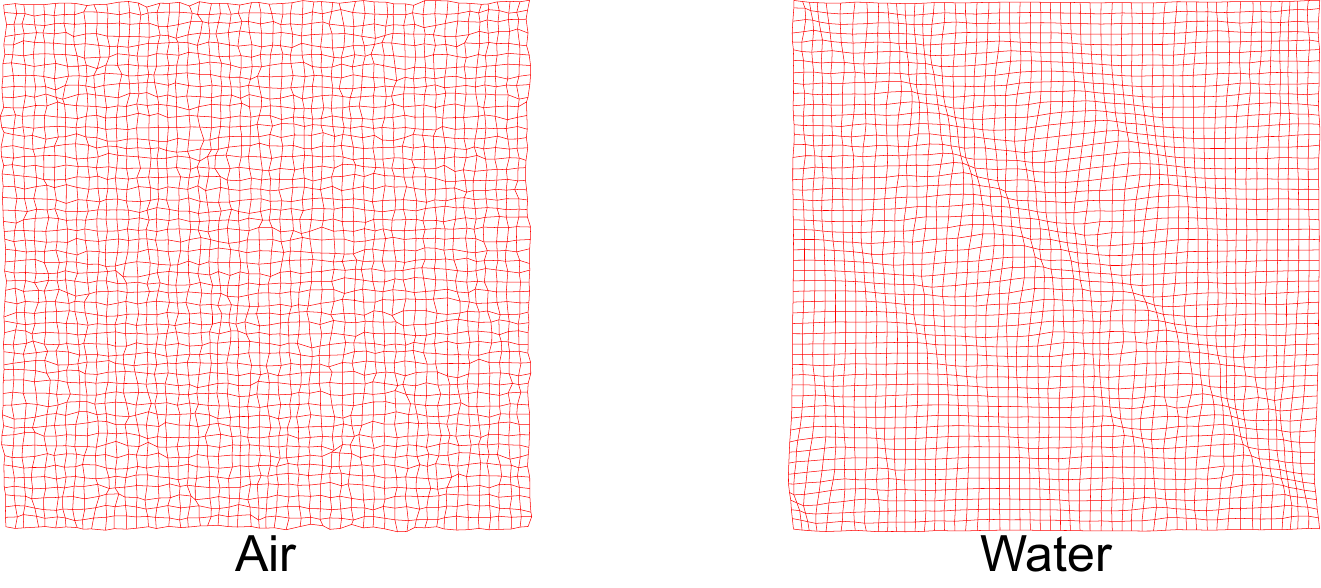}
    \caption{Examples of turbulence fields. The Left shows the turbulence field of the air turbulence. The right shows the turbulence field of the water turbulence.}
    \label{fig:turbulence}
\end{figure}

In this subsection, we present the experimental results of image restoration for turbulence-distorted images using our proposed model, as introduced in subsection \ref{imagerestoration}. Turbulence-distorted images commonly occur when imaging through turbulent refractive media, such as air and water, owing to the refraction and scattering of light \cite{li2021unsupervised}. These distortions pose significant challenges in achieving high-quality and undistorted images for further analysis.

\begin{figure*}[t]
    \centering
    \includegraphics[width=\textwidth]{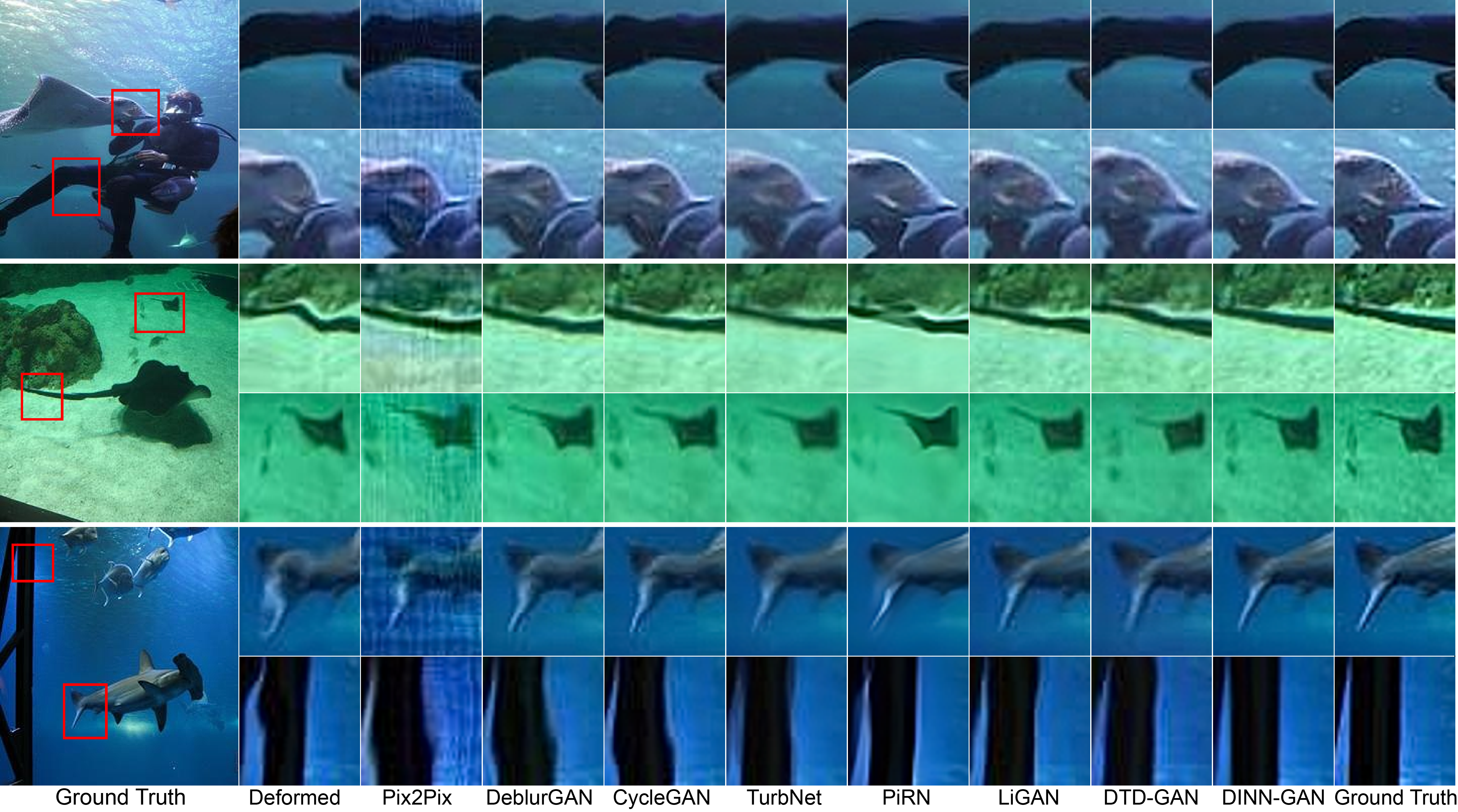}
    \caption{Results of image restoration for images corrupted by ocean type water turbulence by different methods.}
    \label{fig:imagenet_ocean}
\end{figure*}

In our experiments, we obtained the training dataset through the following process. To simulate air-turbulence distortion in images, we utilized the model proposed by \cite{schwartzman2017turbulence}. This model requires specific parameters related to the virtual camera. For our experiment, we set the focal distance of the virtual camera to $300mm$, with a lens diameter of $5.357cm$ and a pixel size of $4\times 10^3mm$. The virtual camera was positioned at an elevation of $4m$ with an object distance of $2km$. For weak turbulence, we set the turbulence strength parameter $C^2_n = 3.6 \times 10^{-13}$, and for strong turbulence, we used $C^2_n = 3.6 \times 10^{-12}$. By applying the turbulence fields obtained from the simulated turbulence model, we distorted the IMAGENET dataset, resulting in a training dataset comprising air-turbulence distorted images. To create the water turbulence image dataset, we utilized a physics-based ray tracer following the methodology described in \cite{thapa2020dynamic}. This approach allowed us to simulate various types of waves for realistic water deformations. In our experiment, we specifically employed two water deformation types: \textit{Ripple} and \textit{Ocean}. Some examples of the turbulence fields are shown in Figure \ref{fig:turbulence}.

\begin{figure}[t]
    \centering
    \includegraphics[width=\textwidth]{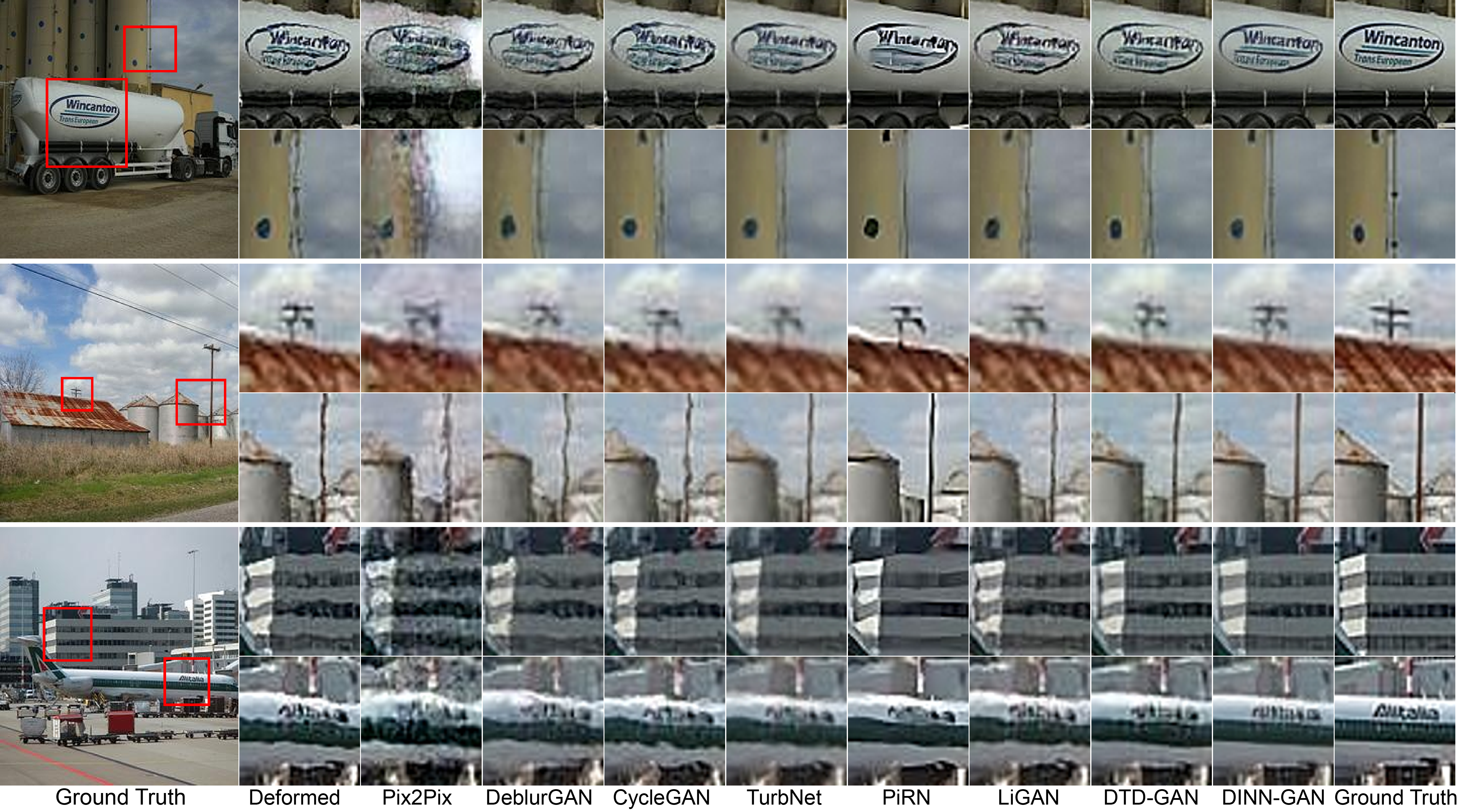}
    \caption{Image restoration results for images distorted by air turbulence by different methods.}
    \label{fig:imagenet_air}
\end{figure}
\begin{landscape}
\begin{table}
\centering
\scriptsize
\begin{tabular}{c|c|cccccccc}
\toprule
Turb. Type             & {Metrics}      & Pix2Pix   & DeblurGAN & CycleGAN  & TurbNet   & PiRN    & LiGAN   & DTDGAN    & DINN-GAN \\\hline
\multirow{3}{*}{Ripple}     & PSNR      & 19.0876   & 20.9176   & 20.9837   & 23.2751   & 24.1916   & 24.8673   & 24.9724   & \textbf{25.3161}\\
                            & SSIM      &  0.4184   &  0.5452   &  0.5376   &  0.6794   &  0.7126   &  0.7351   &  0.7538   &  \textbf{0.8122}\\
                            & MSE       &  0.0294   &  0.0327   &  0.0319   &  0.0213   &  0.0171   &  0.0137   &  0.0133   &  \textbf{0.0127} \\\hline
\multirow{3}{*}{Ocean}      & PSNR      & 19.1837   & 20.8915   & 21.2184   & 23.0734   & 23.3158   & 24.3584   & 24.8440   & \textbf{25.2048}\\
                            & SSIM      &  0.4158   &  0.5363   &  0.5430   &  0.6538   &  0.6841   &  0.7214   &  0.7596   &  \textbf{0.8060}\\
                            & MSE       &  0.0286   &  0.0334   &  0.0327   &  0.0231   &  0.0204   &  0.0168   &  0.0139   &  \textbf{0.0121} \\\hline
\multirow{3}{*}{Air (Weak)} & PSNR      & 20.3745   & 20.1546   & 20.4157   & 22.2937   & 22.8305   & 22.1735   & 22.6834   & \textbf{23.0725}\\
                            & SSIM      &  0.4275   &  0.5175   &  0.5294   &  0.5918   &  0.6315   &  0.6175   &  0.6237   &  \textbf{0.6427}\\
                            & MSE       &  0.0252   &  0.0317   &  0.0292   &  0.0254   &  0.0204   &  0.0267   &  0.0255   &  \textbf{0.0191} \\\hline
\multirow{3}{*}{Air (Strong)}&PSNR      & 18.6375   & 18.6767   & 19.2431   & 20.4266   & 21.9627   & 21.2709   & 21.5145   & \textbf{22.1594}\\
                            & SSIM      &  0.4071   &  0.4552   &  0.4904   &  0.5316   &  0.6092   &  0.5968   &  0.6018   &  \textbf{0.6277}\\
                            & MSE       &  0.0278   &  0.0341   &  0.0322   &  0.0293   &  0.0250   &  0.0287   &  0.0274   &  \textbf{0.0221} \\\hline
\bottomrule
\end{tabular}
\caption{Quantitative comparison of image restoration results for images corrupted by air and water turbulence using various methods.}
\label{tb:restoration}
\end{table}
\end{landscape}

In all of our experiments, we utilized a total of $360,000$ images for training purposes. Additionally, we reserved a separate set of $40,000$ images specifically for testing and evaluation. 

We compare our proposed DINN-GAN model with other state-of-the-art methods for the restoration of distorted images, namely Pix2Pix\cite{van2016conditional}, DeblurGAN\cite{kupyn2018deblurgan}, CycleGAN\cite{zhu2017unpaired}, TurbNet\cite{mao2022single}, \han{PiRN\cite{jaiswal2023physics}}, LiGAN\cite{li2018learning}, and DTD-GAN\cite{rai2022removing}. Figure \ref{fig:imagenet_ocean} presents the image restoration results for images distorted by ocean turbulence. Our proposed method achieves the best results by successfully removing geometric distortions. In contrast, the restored images produced by other methods still exhibit some turbulence distortions. Similarly, Figure \ref{fig:imagenet_air} displays the image restoration results for images distorted by air turbulence, where, once again, our method outperforms the other approaches. Table \ref{tb:restoration} provides a quantitative comparison among the different methods, further demonstrating that our method delivers the best results. \han{Based on the Denoising Diffusion Probability Model, PiRN obtained visually cleaner images due to the denoising processes. However, due to the lack of geometric correction, it can not geometrically restore the distorted images well.}

\begin{figure}
    \centering
    \includegraphics[width=0.9\textwidth]{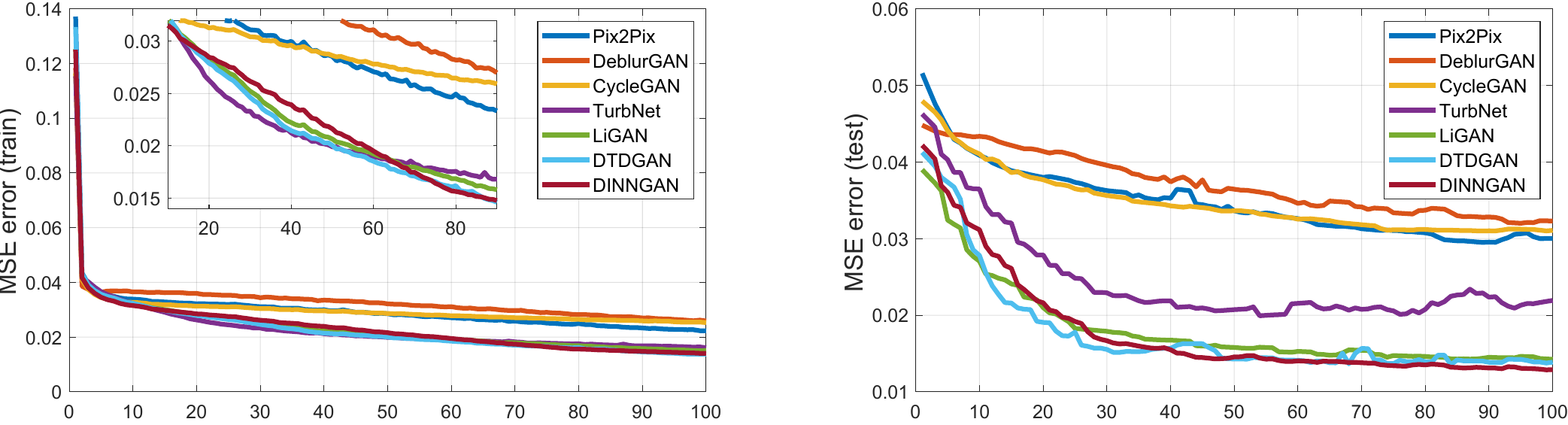}
    \caption{Error plot on the training and testing dataset for turbulence removal experiment for ocean-type turbulence.}
    \label{fig:imagenet_ocean_error}
\end{figure}

Figure \ref{fig:imagenet_ocean_error} illustrates the MSE error \han{in the first $100$ epochs for turbulence removal experiment for ocean-type turbulence over the training and testing, respectively. Note that only the GAN-based models are compared in this figure, as other methods require a different scale of training epochs.} These plots provide valuable insights into the convergence behavior of our DINN model. From the left plot for the MSE error on the training dataset, our model exhibits a slightly slower decrease in training loss compared to LiGAN and DTDGAN. This can be attributed to the constraint imposed by quasiconformal mappings. However, in the right plot on the testing dataset, our method achieves a smaller MSE error on the testing loss.

\begin{figure*}[t]
    \centering
    \includegraphics[width=0.8\textwidth]{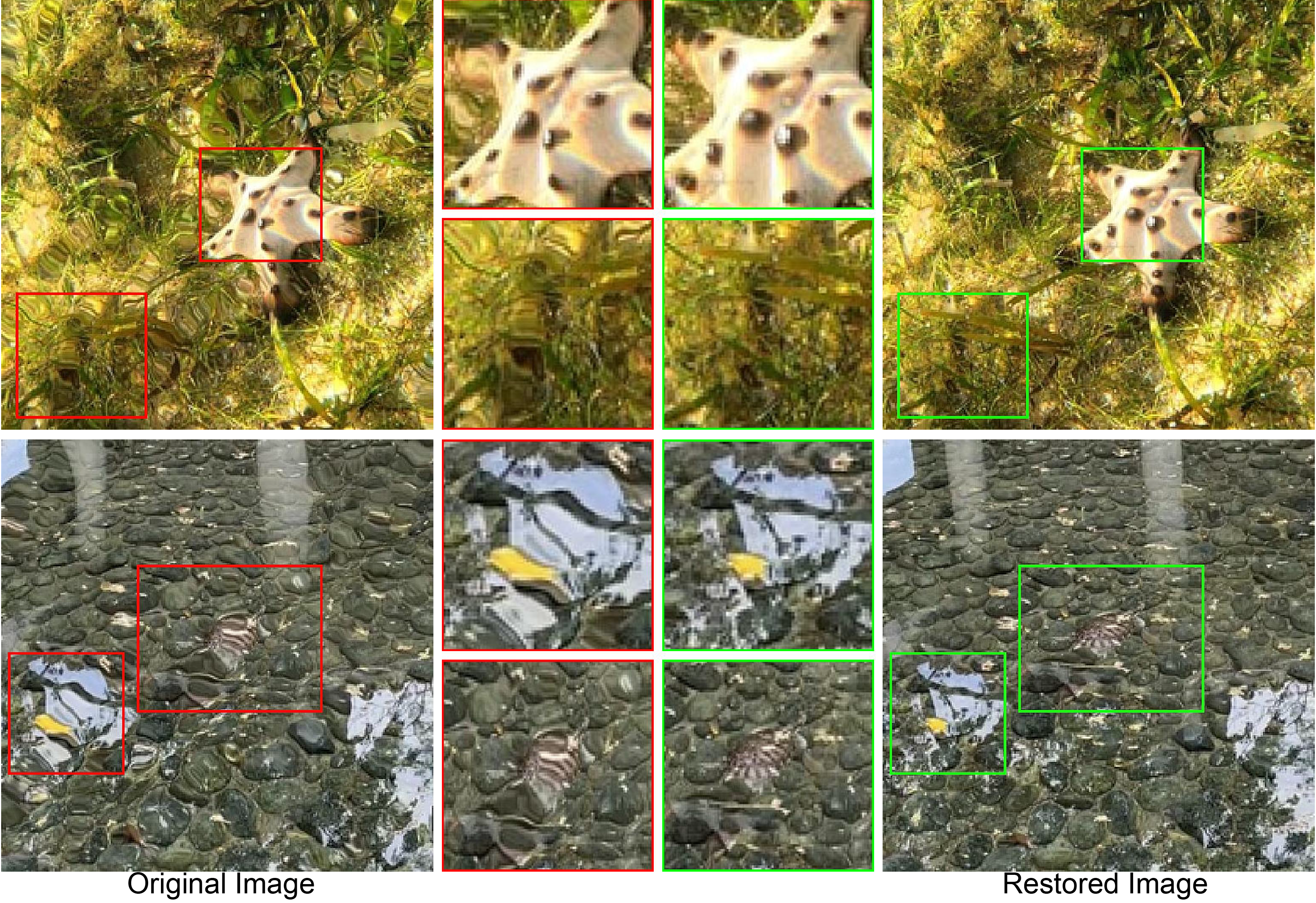}
    \caption{Image restoration results for real images distorted by water turbulence.}
    \label{fig:real_water}
\end{figure*}

Furthermore, we assessed the performance of the DINN-GAN model on real images captured by a digital camera, capturing scenes inside a pool where the images were distorted due to water flow. The distorted images are presented in Figure \ref{fig:real_water} (left), while Figure \ref{fig:real_water} (right) displays the restored image using the DINN-GAN model. Our model adeptly restores the images by effectively removing the geometric distortions. This once again highlights the effectiveness of our proposed method.

\subsection{Distorted Facial Recognition}
\begin{figure*}[t]
    \centering
    \includegraphics[width=\textwidth]{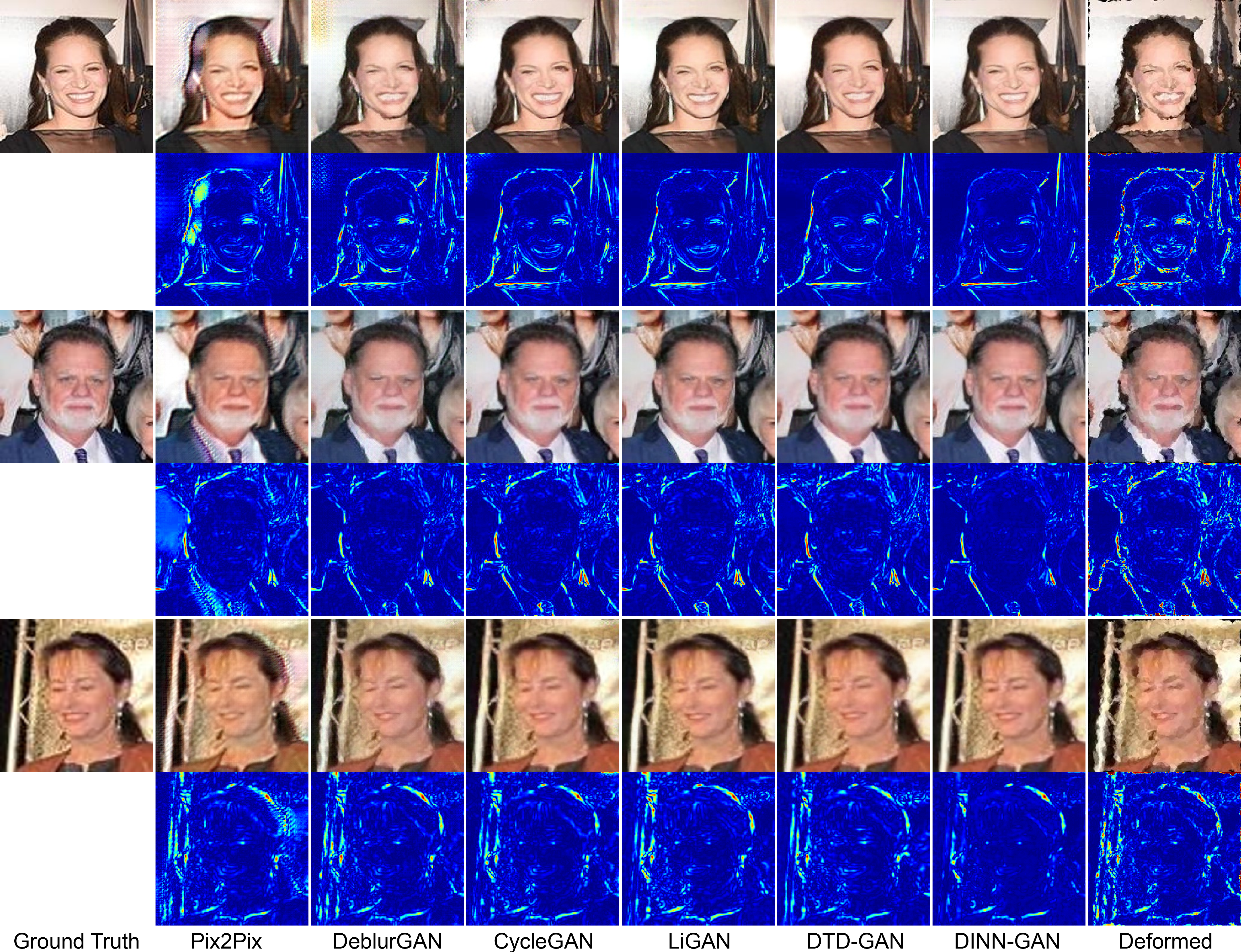}
    \caption{Image restoration results for images severely distorted by strong air turbulence using different methods. The corresponding error maps, defined as the mean square error between the predicted and ground truth values at each pixel point, are shown below the restored images. In the error maps, red indicates higher error, while blue represents lower error.}
    \label{fig:vggface}
\end{figure*}
\begin{table*}[h]
\centering
\scriptsize
\begin{tabular}{c|ccccccccc}
\toprule
Metrics     & Sharp     & Distorted & Pix2Pix   & DeblurGAN & CycleGAN  & LiGAN  & DTDGAN & DINN-GAN\\\hline
PSNR        & $/$       & 21.6473   & 22.1846   & 24.4311   & 24.3958   & 25.3578   & 25.5666   & \textbf{26.3400}\\
SSIM        & $/$       & 0.6738    & 0.6185    & 0.7126    & 0.7129    & 0.7931    & 0.8015    & \textbf{0.8604}\\
MSE         & $/$       & 0.0097    & 0.0077    & 0.0081    & 0.0084    & 0.0067    & 0.0065    & \textbf{0.0063}\\
Accuracy    & 95.31     & 81.23     & 83.58     & 85.08     & 84.98     & 86.76     & 88.53     & \textbf{90.15}\\\hline
\bottomrule
\end{tabular}
\caption{Evaluation of recognition accuracies and image qualities for distorted human faces using different methods.}
\label{tb:vggface}
\end{table*}

In this subsection, we assessed the performance of our proposed model for 1-1 facial verification, as described in Section \ref{1-1facialverification}. We evaluated our method on images distorted by strong air turbulence, which introduces significant geometric distortions that pose a substantial challenge for accurate facial recognition. Similar to the previous subsection, we employed the model proposed by \cite{schwartzman2017turbulence} to generate the dataset. This experiment used the same set of parameters as in the previous subsection, with the turbulence strength parameter set to $C^2_n = 3.6 \times 10^{-13}$. We utilized 450 subjects for training and reserved the remaining 50 subjects for testing.

Figure \ref{fig:vggface} showcases some ground truth facial images in the first column, while the last column displays the same images distorted by strong air turbulence. The figure also presents the restored images produced by our proposed method, along with p2pGAN, DeblurGAN, CycleGAN, LiGAN, and DTD-GAN. Additionally, for each example, corresponding error maps are included, which are defined as the mean square error between the predicted and ground truth values at each pixels. In these error maps, red indicates higher error, while blue represents lower error. Evidently, the restored images using the DINN framework exhibit the most favorable results. This observation is further supported by the quantitative comparison presented in Table \ref{tb:vggface}, which also includes the accuracy of 1-1 facial verification using different methods. Once again, our method achieves significantly higher accuracy compared to the other methods.

\subsection{Self Ablation}

\begin{table}[!h]
\centering
\begin{tabular}{c|cccc}
\toprule
Depth of Net    & 2         & 3         & 4         & 5\\\hline
PSNR            & 20.2672   & 21.4485   & 21.6674   & 21.6931\\
SSIM            &  0.5325   &  0.5977   & 0.6134    & 0.6348\\
MSE             &  0.0339   &  0.0294   & 0.0281    & 0.273\\\hline
\hline
Conv. Type  & Quadruple & Triple    & Double    & Single   \\\hline
PSNR        & 21.8411   & 21.7485   & 21.4485   & 20.0812\\
SSIM        &  0.6353   &  0.6163   &  0.5977   &  0.5184\\
MSE         &  0.0252   &  0.0268   &  0.0294   &  0.0326\\\hline
\bottomrule
\end{tabular}
\caption{Quantiative measurements of the self-ablation tests.}
\label{tb:convdepth}
\end{table}

The performance of the proposed Deformation-Invariant Neural Network (DINN) as a learning approach may rely on the chosen network architecture. It is crucial to strike a balance between the depth of the architecture and its convergence capabilities. An architecture that is too shallow may fail to converge adequately, resulting in suboptimal performance. On the other hand, an excessively deep architecture may lead to excessive training consumption. To determine the optimal configuration, we conduct two self-ablation studies within our model. Specifically, we investigate the influence of various downsample levels and convolution layers on the performance of the DINN-based image restoration model. This analysis allows us to gain insights into the impact of these factors on the overall effectiveness of the DINN framework.

\textbf{Influence of Downsample Levels}
The downsampled levels in the encoder-decoder architecture, utilized by the BC estimator, play a vital role in our model. To assess the influence of the number of downsampling levels on the generalization ability of our model, we conducted a self-ablation study. The results presented in Table \ref{tb:convdepth} compellingly demonstrate that a UNet architecture with a downsample level of 3 is sufficient to achieve effective learning and generate high-quality restoration mappings. This finding highlights the importance of striking a balance between the complexity of the architecture and its performance.

\medskip

\textbf{Influence of Convolution Layers}

The number of convolution layers in each level of the encoder-decoder architecture is another crucial factor that affects the restoration quality. The results presented in Table \ref{tb:convdepth} indicate that employing a double-convolution configuration for each level in the encoder-decoder architecture is highly effective in producing satisfactory restoration results. This finding highlights the importance of carefully considering the number of convolution layers at each level to achieve optimal performance in the restoration process.

\section{Conclusion and Future Work}
In this paper, we have introduced the deformation-invariant neural network (DINN) framework to solve the challenging problem of imaging tasks involving geometrically distorted images. Our proposed framework, incorporating the quasiconformal transformer network (QCTN), has demonstrated its effectiveness in addressing various imaging tasks, including image classification of distorted images, image restoration in the presence of atmospheric or water turbulence and 1-1 facial verification under strong air turbulence.

The key contributions of our work include the development of DINN, which ensures consistent latent features for geometrically distorted images capturing the same underlying object or scene. We have introduced the portable QCTN component, which allows large pretrained networks to process heavily distorted images without requiring additional tuning, thereby reducing computational costs. The QCTN generates bijective deformation maps that preserve the salient features of the original images, resulting in more accurate restoration and recognition results. Our experimental results have shown that the proposed DINN framework outperforms existing GAN-based restoration methods in scenarios involving atmospheric turbulence and water turbulence. Furthermore, the application of DINN to 1-1 facial verification under strong air turbulence has demonstrated its efficacy in enhancing the accuracy of facial recognition even in adverse conditions.

While our proposed framework has yielded promising results, there are still several avenues for future research. One potential direction is to investigate the application of the DINN framework to other imaging tasks, such as image registration and image segmentation. Additionally, it is worth noting that the current DINN framework may yield less satisfactory outcomes when confronted with very extreme deformations. Therefore, further exploration is needed to enhance the ability of the proposed model to handle such challenging scenarios.

In conclusion, the proposed DINN framework, incorporating the QCTN component, offers a powerful solution for addressing imaging tasks involving geometrically distorted images. Our experimental results have demonstrated its superiority in image classification, image restoration and facial verification tasks under challenging conditions. The DINN framework opens up new possibilities for handling geometric distortions in various applications and provides a valuable contribution to the field of deep learning in imaging and computer vision.

\section*{Acknowledgment}
Lok Ming Lui was supported by HKRGC GRF (Project ID: 14305919 and 143062721). Thanks to the support of Hong Kong Center for Cerebro-Cardiovascular Health Engineering.
\begin{scriptsize}
\bibliographystyle{elsarticle-num}  
\bibliography{reference}
\end{scriptsize}

% \bigskip
% \setlength\intextsep{0pt} % align top of photo with text
%     \noindent \small \textbf{Han Zhang} is currently a Ph.D student in the Department of Mathematics at City University of Hong Kong. Han’s research interest includes computational geometry, scientific computing and image processing.    
% \bigskip
% \setlength\intextsep{0pt}
%     \noindent \small \textbf{Lok Ming Lui} received his Ph.D. degree in applied mathematics from the University of California at Los Angles in 2008. He is a Full Professor at the Department of Mathematics, Chinese University of Hong Kong. His current research interests include computational conformal and quasiconformal geometry, Teichmuller theory, surface registration, medical imaging, and shape analysis.
\end{document}